\documentclass[runningheads]{llncs}

\usepackage[mobile]{eccv}

\usepackage{eccvabbrv}

\usepackage{graphicx}
\usepackage{booktabs}
\usepackage{multirow}

\usepackage[accsupp]{axessibility}  
\usepackage{tikz}
\usepackage{xcolor}
\definecolor{OursWin}{HTML}{4A90E2}   
\definecolor{BothSame}{HTML}{D3D3D3}  
\definecolor{OursLoss}{HTML}{F5A623}

\newcommand{\TriBar}[3]{%
  \begin{tikzpicture}[baseline=-0.5ex]
    \pgfmathsetmacro{\LWin}{#1 * 2.5}
    \pgfmathsetmacro{\LSame}{#2 * 2.5}
    \pgfmathsetmacro{\LLoss}{#3 * 2.5}
    \fill[OursWin]  (0,0) rectangle (\LWin, 0.3);
    \fill[BothSame] (\LWin,0) rectangle (\LWin+\LSame, 0.3);
    \fill[OursLoss] (\LWin+\LSame,0) rectangle (\LWin+\LSame+\LLoss, 0.3);
  \end{tikzpicture}%
}

\usepackage[pagebackref,breaklinks,colorlinks,citecolor=eccvblue]{hyperref}

\usepackage{orcidlink}

\begin{document}

\title{EMOSH: Expressive Motion and Shape Disentanglement for Human Animation} 

\titlerunning{EMOSH}

\makeatletter
\newcommand{\printfnsymbol}[1]{%
  \textsuperscript{\@fnsymbol{#1}}%
}
\renewcommand*{\@fnsymbol}[1]{\ensuremath{\ifcase#1\or *\or \dagger\or \ddagger\or
   \mathsection\or \mathparagraph\or \|\or **\or \dagger\dagger
   \or \ddagger\ddagger \else\@ctrerr\fi}}
   
\author{Dongbin Zhang$^{1,2*}$ \and
Hao Liu$^{2}$ \and 
Binquan Dai$^{1}$\and
Kangjie Chen$^{1}$\and
Chuming Wang$^{1}$ \and Chen Li$^{2\dagger}$ \and Jing LYU$^{2}$ \and Haoqian Wang$^{1\dagger}$
}

\authorrunning{D.~Zhang et al.}

\institute{$^1$Tsinghua Shenzhen International Graduate School, Tsinghua University \\ $^2$WeChat Vision, Tencent Inc. \\ Project Page: \textcolor{magenta}{https://eastbeanzhang.github.io/EMOSH/}}

{\let\thefootnote\relax\footnotetext{{$^{*}$ Intern at WeChat Vision. ~ $^{\dagger}$ Corresponding authors. }}}

\maketitle

\begin{figure}[h]
  \centering
  \includegraphics[width=\textwidth]{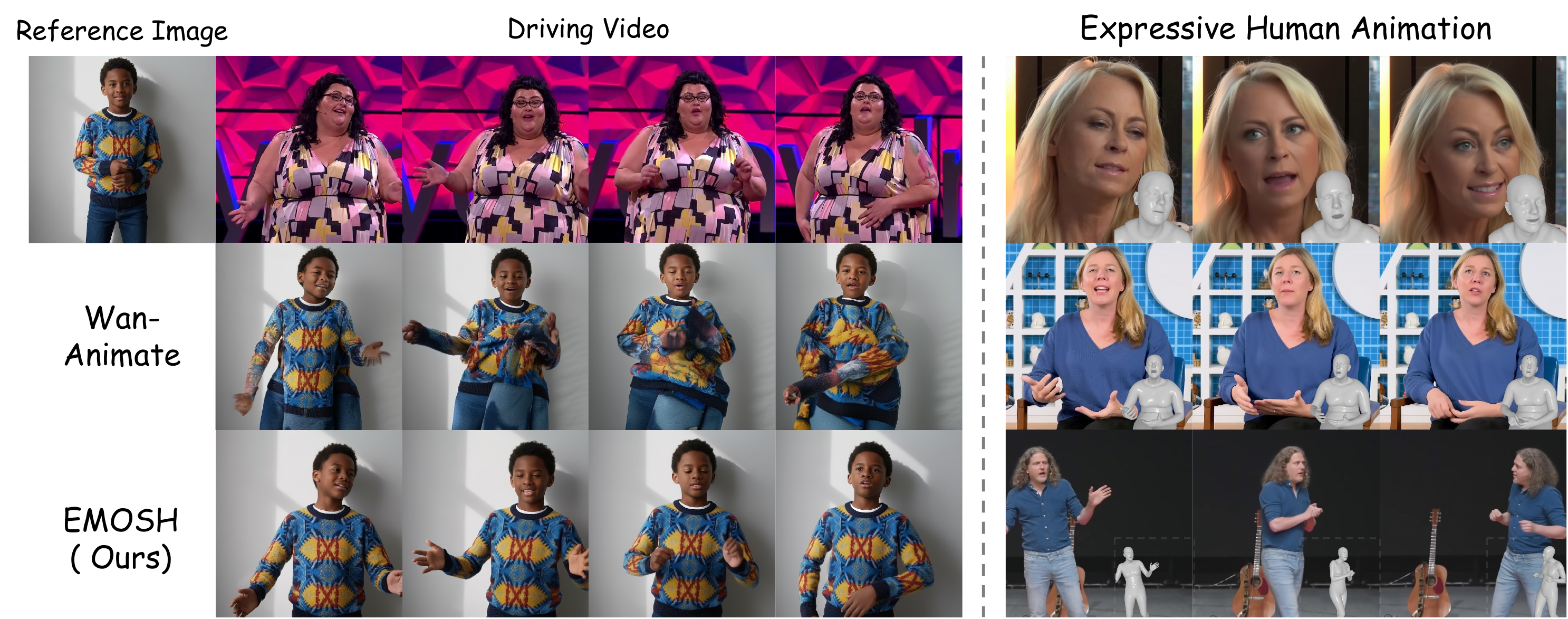}
 \caption{Given a reference image and a driving video, EMOSH achieves high-fidelity, mesh-guided expressive human animation while disentangling expressive motion from body shape to prevent shape leakage.
  }
  \label{fig:teaser}
  \vspace{-28pt}
\end{figure}

\begin{abstract}
High-fidelity and expressive controllable human animation is essential for content creation and digital avatar applications. However, existing methods face a dilemma between expressiveness and disentanglement. Mainstream 2D pose-conditioned approaches suffer from "motion-shape entanglement", leading to the leakage of the driving subject's body shape. Conversely, methods relying on 3D priors (e.g., SMPL) achieve geometric disentanglement but struggle to capture facial expressions and complex gestures, resulting in rigid animations.
To this end, we propose EMOSH, a novel framework for high-fidelity controllable human video generation. First, an Expressive Human Model (EHM) is introduced as the core control representation. By explicitly disentangling shape and pose parameters, we fundamentally resolve the body shape leakage issue. Alongside this, a robust motion tracker is designed to accurately estimate EHM parameters from video. Second, we propose a Coarse-to-Fine Hybrid Motion Injection strategy, enabling more fine-grained control over expressions and gestures. Furthermore, we introduce a Spatially-Aligned Conditioning mechanism to bridge the domain gap between training and inference, improving identity consistency. Extensive experiments demonstrate that EMOSH outperforms previous methods in both self-driven and cross-driven scenarios, producing high-fidelity videos with vivid expressions while maintaining shape disentanglement. 

\end{abstract}
\section{Introduction}
\label{sec:intro}
Driven by rapid advancements in generative AI, high-fidelity and expressive human animation has become a research hotspot in computer vision, with broad applications in digital avatars, film, and social media. The core objective is to animate a reference image using a driving video. This requires faithfully replicating the driver's motions, expressions, and spatial layout, while strictly preserving the reference subject's identity and appearance.

Extensive research has explored various methodologies in this domain. Early works~\cite{fomm,deformable_gan,gac_gan} primarily relied on GANs~\cite{gan} and warping fields, but these methods often fail with large motions or unseen regions. Recently, diffusion models~\cite{ddpm,scoresde} have become the mainstream choice due to their superior generation quality. Pioneering explorations like ControlNet~\cite{controlnet} established the paradigm of injecting 2D skeletal poses as spatial conditions into diffusion models. Building on this, works like Animate Anyone~\cite{animateanyone} introduced temporal layers for more fluid motion transfer, while MagicPose~\cite{magicpose} incorporated fine-grained facial landmarks to enhance expression control. Most recently, driven by Video Foundation models, approaches~\cite{vace,hypermotion,flexiact,x-unimotion,dreamactor-m1} have scaled human animation to large DiT-based backbones, further elevating generation quality and dynamic fidelity.

Despite these advancements, existing methods still face a dilemma between \textbf{expressiveness} and \textbf{disentanglement}. Mainstream approaches relying on 2D poses suffer from severe \textbf{"Motion-Shape Entanglement"}. Because 2D skeletons implicitly encode the driver's body proportions—an issue difficult to resolve even with pose retargeting—these models inevitably leak the driver's physique into the generated output. To address this, some works~\cite{champ,realisdance-dit} adopt 3D parametric models (e.g., SMPL~\cite{smpl}) to geometrically disentangle shape and motion. However, such priors lack the capacity to capture and represent facial expressions and complex gestures, resulting in \textbf{rigid animations}. Furthermore, a pervasive \textbf{domain gap} exists between training and inference. Models are typically trained on spatially aligned self-driven data but must handle spatially misaligned cross-driven tasks during inference. This discrepancy frequently causes identity loss and visual artifacts, particularly in long video generation.

To address these challenges, we propose EMOSH, a novel framework for high-fidelity human video generation that achieves both expressive motion control and shape disentanglement, as shown in \cref{fig:teaser}. First, we adopt a 3D parametric model, the \textbf{Expressive Human Model (EHM)}~\cite{guava}, as our core representation to resolve the shape leakage of 2D pose-based methods. By explicitly disentangling shape and pose parameters, EHM allows us to precisely transfer driving motions while preserving the reference subject's body shape. Alongside this, we design a \textbf{Confidence-Aware Motion Tracker} to robustly capture full-body motions—including facial expressions and hand gestures—from monocular videos. Second, to overcome the lack of high-frequency details (e.g., eye blinking) in naive mesh guidance, we propose a \textbf{Coarse-to-Fine Hybrid Motion Injection strategy}. This combines global geometric priors of 3D meshes with the local precision of 2D keypoints, enhancing control precision while maintaining structural integrity. Finally, to bridge the domain gap between training and inference, we introduce a \textbf{Spatially-Aligned Conditioning mechanism}. By injecting spatially-aligned visual anchors, it mitigates error accumulation, reduces visual artifacts, and ensures superior identity consistency in video generation.

Extensive experiments demonstrate that EMOSH achieves state-of-the-art performance in both self-driven and cross-driven scenarios, surpassing 2D pose-based methods plagued by shape coupling and previous mesh-based methods limited by expressiveness. It successfully achieves high-fidelity motion-shape disentanglement while preserving vivid expressions, gestures, and identity. In summary, our main contributions are as follows:

\begin{itemize}
\item We propose EMOSH, a novel framework that utilizes the Expressive Human Model for controllable human video generation. It delivers highly expressive human animation while strictly maintaining motion-shape disentanglement.

\item We design a Confidence-Aware Motion Tracker to accurately capture expressive motions from videos, coupled with a Coarse-to-Fine Hybrid Motion Injection strategy that facilitates highly precise control.

\item We propose a Spatially-Aligned Conditioning strategy to mitigate the training-inference domain gap. This mechanism reduces visual artifacts and improves identity preservation, setting a new SOTA over existing methods.
\end{itemize}
\section{Related Work}
\label{sec:related}

\subsection{Video Generation}
\label{subsec:vidgen}
Driven by the success of diffusion models~\cite{ddim,ddpm} in image synthesis~\cite{stable_diffusion,dall-e,imagen}, the video generation paradigm has rapidly shifted away from traditional GANs~\cite{mocogan,dvdgan} and autoregressive models~\cite{videogpt,phenaki}. Early video diffusion models~\cite{vdm,make-a-video} typically adapted pre-trained image models by inflating 2D U-Nets into 3D U-Nets and adding temporal layers. To further enhance visual quality, works like Imagen-Video~\cite{imagen_video} introduced cascaded frameworks, initially generating low-resolution videos and progressively upscaling them using spatial super-resolution and temporal interpolation models.

Recently, Diffusion Transformers (DiT)~\cite{dit} have emerged as a powerful alternative to U-Net backbones, overcoming the scalability bottlenecks of convolutional networks and exhibiting strong scaling laws. Leveraging this architecture and massive video datasets, subsequent works~\cite{sora,opensora,runwaygen,seedancev1} have significantly scaled up video generation. These DiT-based models achieve high-quality, high-resolution long video synthesis with remarkable physical realism.

Within the DiT paradigm, various architectural designs have emerged. For instance, HunyuanVideo~\cite{hunyuanvideo} trains a causal 3D VAE for efficient compression and employs a dual-stream DiT to process text and video tokens independently. Notably, it replaces standard factorized spatiotemporal attention with full 3D attention, while extending RoPE~\cite{rope} to 3D to support dynamic resolutions and durations. In contrast, Wan2.1~\cite{wan} introduces Wan-VAE with caching mechanisms to alleviate memory bottlenecks during long video encoding. Structurally, it adopts a single-stream DiT that injects text via cross-attention and utilizes a shared MLP for timestep modulation to improve parameter efficiency.

Concurrently, numerous studies are dedicated to advancing specific dimensions of video generation, including consistent generation~\cite{genie3,voyager,gamecraft}, acceleration~\cite{flashvideo,histream}, duration extension~\cite{selfforcing,deepforcing}, and joint audio-video generation~\cite{veo3,ovi}.

\subsection{Human Animation}
\label{subsec:humanani}
Beyond general text-to-video, researchers have actively explored controllable human animation. Early works~\cite{pg2,patn} treated pose transfer as an image-to-image translation task, encoding the reference image into latent features and decoding them after injecting the target pose. However, these methods often struggled with large motions. Subsequently, GAN-based warping techniques became prominent for both facial animation~\cite{face-vid2vid,depth-aware-gen-talk,geometry_contrastive_gan} and body pose transfer~\cite{human_motion_transfer_3d,warping-gan,gac_gan,liquid_warping_gan}. By predicting flow-based warping fields, these approaches spatially transform reference features to align with target poses. While effective at preserving texture details, they are prone to artifacts in occluded regions.

Leveraging the generative power of diffusion models, numerous works~\cite{follow_pose,magicanimate,disco} have adopted ControlNet-like frameworks~\cite{controlnet} to achieve controllable animation driven by 2D poses~\cite{dwpose,openpose}. Animate Anyone~\cite{animateanyone} pioneered the use of ReferenceNet to extract spatial details, injecting them into the main U-Net via attention to improve appearance preservation. In contrast, UniAnimate~\cite{unianimate} simplified this by stacking the reference latent with noisy latents and utilizing Temporal Mamba for efficient long-sequence generation. To mitigate identity loss under extreme motions, StableAnimator~\cite{stableanimator} introduced a distribution-aware ID adapter, while MimicMotion~\cite{mimicmotion} incorporated confidence-aware guidance to handle pose estimation errors. Similarly, HyperMotion~\cite{hypermotion} leverages a DiT-based framework with enhanced RoPE to further improve generation quality. Most recently, Wan-Animate~\cite{wan-animate} unified character animation on the Wan-14B~\cite{wan} foundation, employing implicit encoding to replicate subtle expressions.

Despite significant progress, 2D Pose-based methods remain limited by motion-shape entanglement. To resolve this, several approaches~\cite{champ,human4dit,realisdance,uni3c} employ SMPL-rendered~\cite{smpl} normal or depth maps as conditions. However, these explicit priors are often confined to coarse limb movements and fail to match the fine-grained control over expressions and gestures offered by 2D Pose. In contrast, our EMOSH extends the representation and controllability of 3D mesh, achieving both shape disentanglement and superior motion expressiveness.

Alternatively, methods based on 3D radiance fields (e.g., NeRF~\cite{nerf}, 3DGS~\cite{3dgs}) inherently maintain 3D consistency by driving reconstructed avatars. While offering superior efficiency, approaches based on per-subject optimization~\cite{hravatar,gaussianavatars} or feed-forward reconstruction~\cite{guava,lhm} often struggle to capture complex backgrounds and fine-grained dynamics, such as hair or clothing wrinkles. Consequently, they fall short of the visual realism provided by diffusion-based models.

\section{Method}
\label{sec:method}

\begin{figure}[tb]
  \centering
  \includegraphics[width=\textwidth]{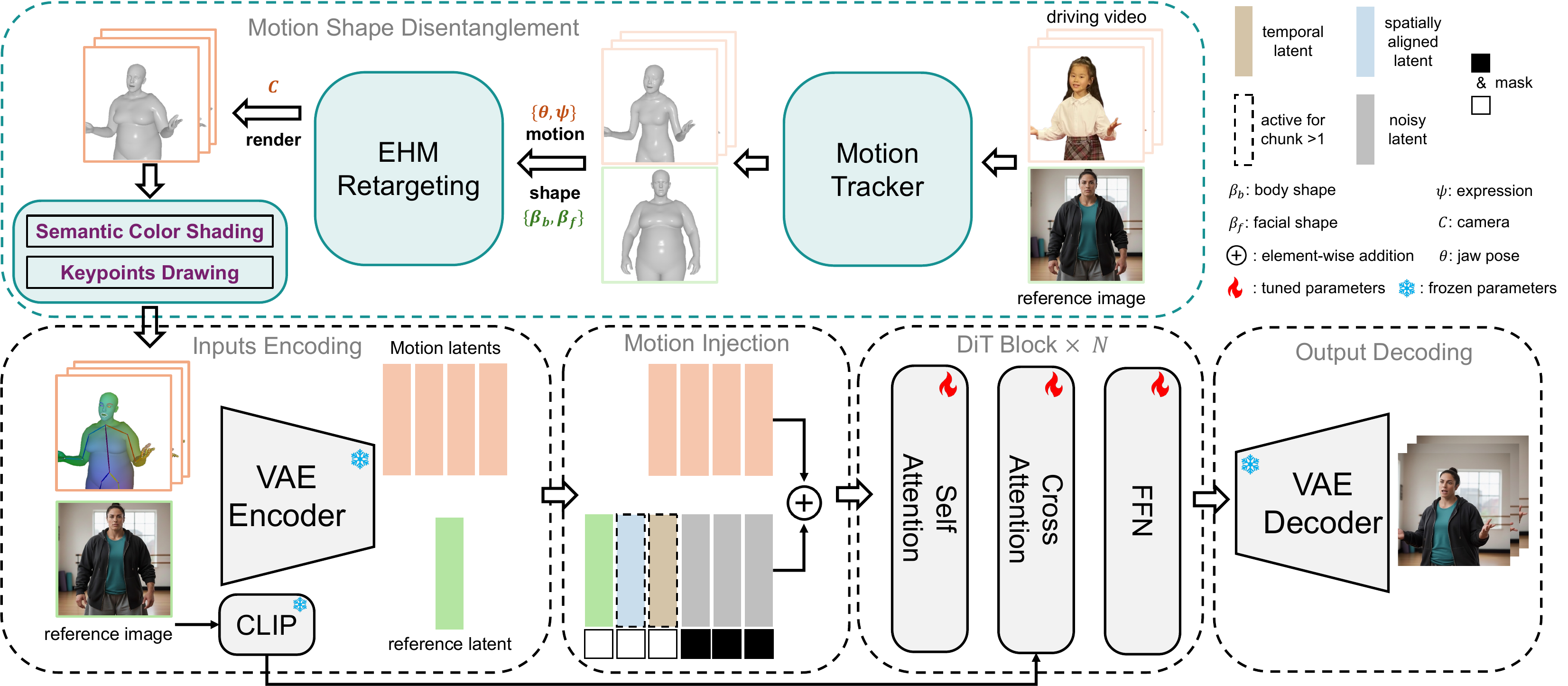}
  \caption{
  First, the motion tracker extracts motion ($\theta^d, \psi^d$) and  camera ($C^d$) parameters from the driving video and shape parameters ($\beta_b^r, \beta_f^r$) from the reference image, achieving motion-shape disentanglement via EHM retargeting. The retargeted model is then rendered into hybrid control signals through semantic color shading and keypoint drawing, and encoded into motion latents. During the generation phase, the reference latent and noisy latent are concatenated; for subsequent video chunks (chunk > 1), the spatially-aligned and temporal latents are additionally activated and concatenated. The motion latents are injected into the main sequence via addition and fed into the DiT network for denoising, and are finally decoded into the output video.
  }
  \label{fig:pipeline}
\end{figure}

As illustrated in \cref{fig:pipeline}, we present the overview of EMOSH. We first review the preliminaries in \cref{subsec:preliminaries}. In \cref{subsec:motion_tracking}, we introduce the Expressive Human Model (EHM) as our motion representation, alongside the design of a robust motion tracker. \cref{subsec:motion_control} elaborates on the Coarse-to-Fine Hybrid Motion Injection strategy, as well as our approach to disentangling motion from body shape. Finally, \cref{subsec:infer_train} describes the Spatially-Aligned Conditioning mechanism designed to enhance identity consistency, and details our training strategy.

\subsection{Preliminaries}
\label{subsec:preliminaries}
\noindent \textbf{Video Generation.} We adopt advanced Wan2.1-I2V~\cite{wan} as our baseline framework. It employs a Diffusion Transformer (DiT)~\cite{dit} to process spatiotemporal latents compressed by a 3D Causal VAE. Text and image semantics are extracted via T5~\cite{t5-encoder} and CLIP~\cite{clip}, respectively, and injected into the backbone through cross-attention. To achieve long video generation, Wan2.1-I2V relies on an explicit latent conditioning strategy: the encoded reference image is concatenated with the noisy latent along the temporal dimension,  while a binary mask is concatenated channel-wise to distinguish reference frames (value 1) from target generation regions (value 0).


\noindent \textbf{Expressive Human Model (EHM).} To address the limited facial expressiveness and head-body shape coupling in SMPL-X~\cite{smplx}, we adopt EHM~\cite{guava}, a mesh-based parametric model integrating SMPL-X with FLAME~\cite{flame}. It represents variations in human shape and expression within a linear space, explicitly decoupling identity into body shape $\beta_b \in \mathbb{R}^{|\beta_b|}$ and head shape $\beta_f \in \mathbb{R}^{|\beta_f|}$, while capturing facial dynamics via expression parameters $\psi \in \mathbb{R}^{|\psi|}$. Skeletal poses—encompassing body, hands, and facial joints—are defined by axis-angle rotations $\theta \in \mathbb{R}^{|\theta|}$. Using Linear Blend Skinning (LBS) for vertex deformation, EHM maps these parameters to a 3D mesh: $V=M_{ehm}(\beta_b, \beta_f, \psi, \theta)$, where $V \in \mathbb{R}^{N \times 3}$ denotes the deformed vertex coordinates.

\subsection{Expressive Motion Representation}
\label{subsec:motion_tracking}
Unlike previous methods~\cite{champ,realisdance} restricted to coarse SMPL-based limb control~\cite{smpl}, we introduce the highly expressive EHM alongside a robust motion tracker. This enables the accurate capture of facial expressions and hand poses, pushing the upper limits of mesh-based control to achieve both rich details and high-fidelity video generation.

Given a driving video, our goal is to estimate the per-frame EHM parameters $\Theta_{ehm}=\{\beta_b,\beta_f,\theta,\psi\}$ and camera parameter $C$. While GUAVA~\cite{guava} offers a preliminary tracking workflow, its cascaded strategy—optimizing facial parameters before body pose—suffers from inefficiency and limitations in challenging scenarios, such as side or back views.

\begin{figure}[tb]
  \centering
  \begin{subfigure}[b]{0.49\textwidth}
    \centering
    \includegraphics[width=\textwidth]{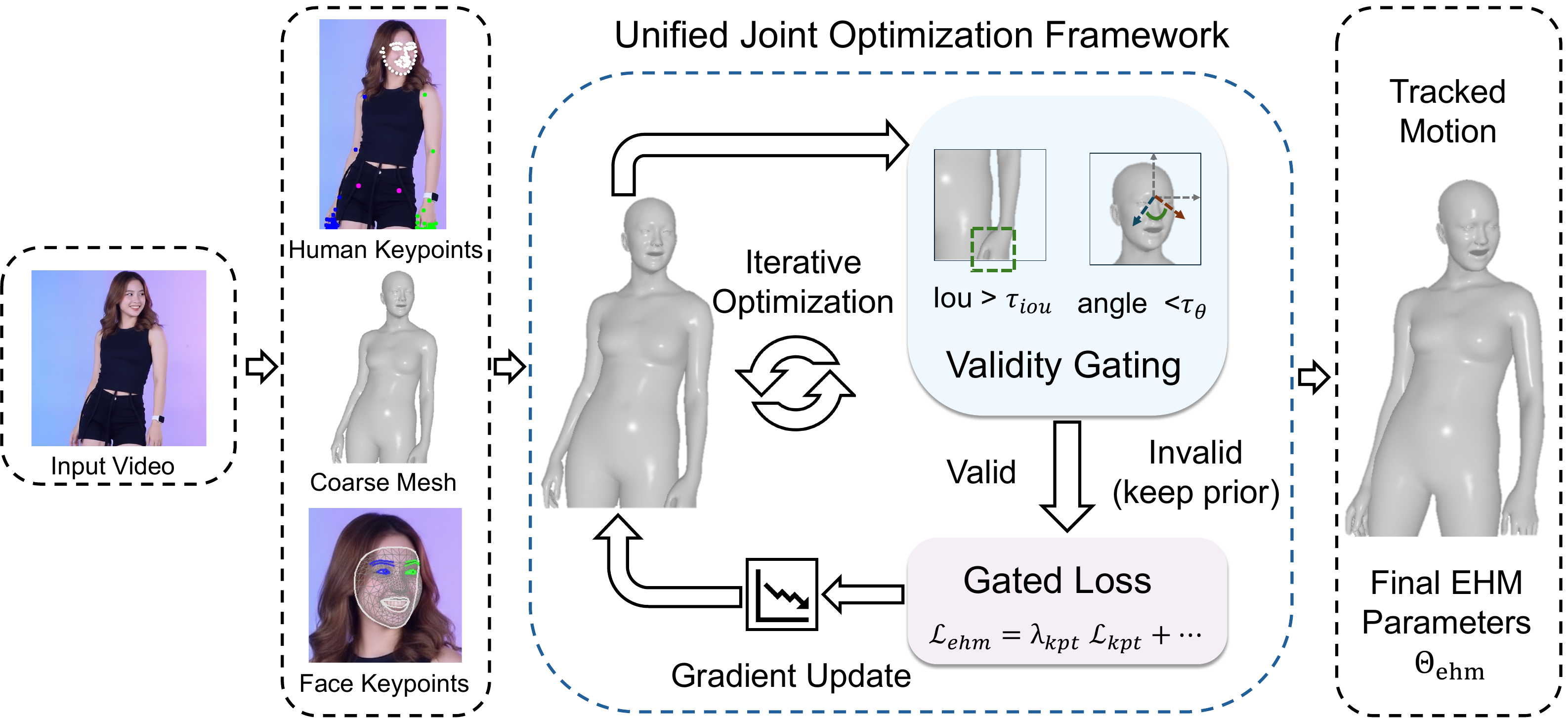}
    \caption{Optimization pipeline of Motion Tracker.}
    \label{fig:motion_tracker}
  \end{subfigure}
  \hfill
  \begin{subfigure}[b]{0.50\textwidth}
    \centering
    \includegraphics[width=\textwidth]{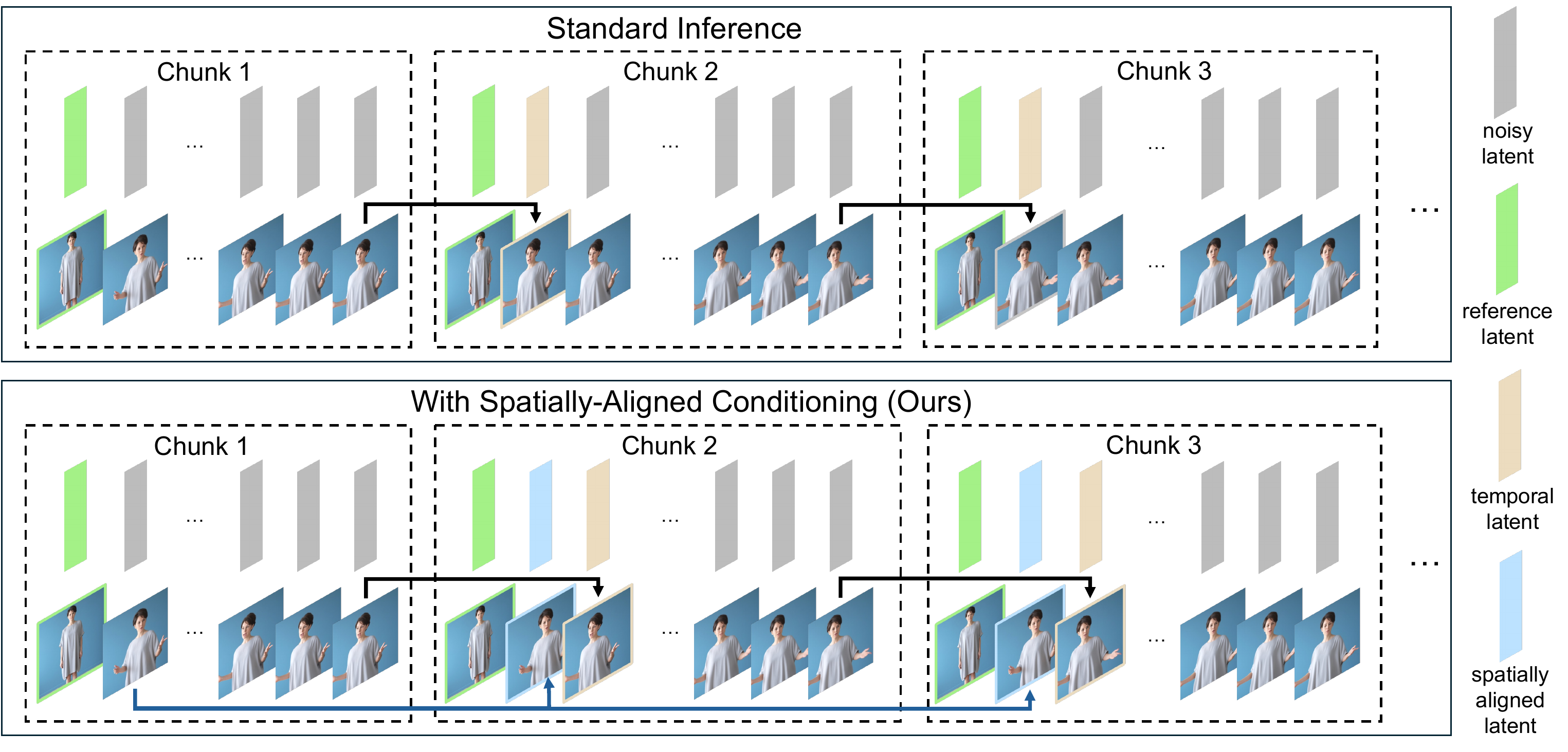}
    \caption{Spatially-Aligned Conditioning strategy.}
    \label{fig:SAC}
  \end{subfigure}
  \caption{(a) It extracts 2D/3D priors from the input video and dynamically filters unreliable guidance signals via Validity Gating, obtaining EHM parameters through joint iterative optimization. (b) For subsequent chunks in long video inference, the initial latent from the first chunk is injected as an additional spatially-aligned latent.}
  \label{fig:tracker_and_SAC}
\end{figure}

As illustrated in \cref{fig:motion_tracker}, we propose a unified joint optimization framework that simultaneously recovers body, hand, facial, and camera parameters, simplifying the pipeline and improving tracking efficiency. The process follows a coarse-to-fine strategy: we first initialize the shape, pose, and camera parameters for each frame using off-the-shelf estimators. Then, leveraging differentiable rendering, we jointly fine-tune these parameters by minimizing the reprojection error $\mathcal{L}_{kpt}$ between the projected EHM vertices and detected 2D keypoints.

\begin{figure}[tb]
  \centering
  \includegraphics[width=\textwidth]{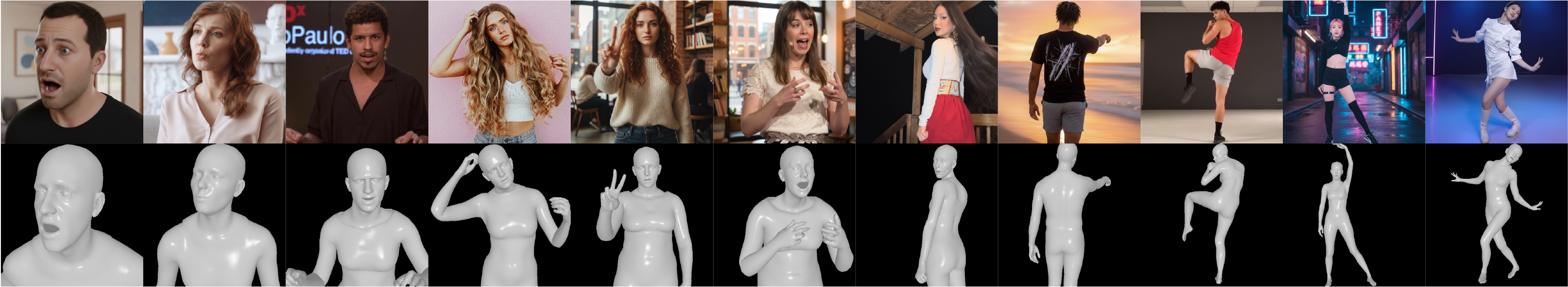}
  \caption{
Visual results of our motion tracker. Our tracker accurately extracts expressive motion parameters across diverse scenarios.
  }
  \label{fig:show_tracking}
\end{figure}

To address the instability of keypoint estimation and enhance robustness under complex motions (e.g., large-angle head turns, limb occlusions), we introduce a Confidence-Aware Validity Gating mechanism. Specifically, for hand tracking, we employ a dual-criteria check based on keypoint confidence and the projection IoU of the initial 3D mesh. The optimizer utilizes keypoint guidance only when hands are strictly visible; otherwise, it falls back to initial priors. For facial tracking, we compute the angle between the head orientation and camera view, dynamically masking the facial keypoint loss during large-angle or back-view scenarios to prevent structural collapse from forced alignment. Finally, incorporating the inter-frame smoothness $\mathcal{L}_{smooth}$, parameter regularization $\mathcal{L}_{reg}$, and 3D hand alignment $\mathcal{L}_{hand}^{3D}$, the overall joint optimization objective is defined as:
\begin{equation}
\mathcal{L}_{ehm} = \lambda_{kpt} \mathcal{L}_{kpt} + \lambda_{3d} \mathcal{L}_{hand}^{3D} + \lambda_{reg} \mathcal{L}_{reg} + \lambda_{smooth} \mathcal{L}_{smooth}.
\end{equation}
As demonstrated in \cref{fig:show_tracking}, our motion tracker enables accurate and stable motion capture across diverse and challenging scenarios.
\subsection{Motion-Appearance Control}
\label{subsec:motion_control}
\noindent \textbf{Hybrid Motion Injection}. To achieve precise motion control, we design a Coarse-to-Fine Hybrid Motion Injection strategy based on the per-frame tracked EHM mesh.
First, to enable the network to spatially distinguish different body regions, we employ a geometric semantic coloring scheme. Specifically, the normalized 3D coordinates of each vertex in a canonical T-pose are mapped to fixed RGB values $V_{color}$. Using a differentiable renderer $\mathcal{R}$, we then render these semantically colored meshes to generate dense 2D condition map for each frame.

However, while this rendering effectively represents macroscopic body structures, it provides predominantly low-frequency signals. In distant views or at low resolutions, the rendered maps often miss subtle high-frequency dynamics, such as mouth closures or blinking. To address this, we introduce sparse 2D keypoints to provide complementary high-frequency details. Specifically, we select a set of key semantic vertices $V_{kpt}$ from the mesh and explicitly draw them onto the rendered map using distinct colors, allowing the model to perceive these detailed changes. The final condition map $I_{mesh}$ is thus formulated as:
\begin{equation}
I_{mesh} = \mathcal{P}\Big(\mathcal{R}(M_{ehm}, V_{color}, C), V_{kpt}\Big)~,
\label{eq:mesh_render}
\end{equation}
where $\mathcal{P}$ denotes the keypoint drawing operation. Finally, $I_{mesh}$ is processed by a VAE encoder and a lightweight embedding layer. The extracted features are directly added to the noisy video latent, providing explicit motion guidance during the denoising process.

\noindent \textbf{Identity Injection.} To effectively inject appearance information from the reference image, we follow the explicit latent conditioning of Wan2.1-I2V~\cite{wan}. Specifically, the VAE-encoded reference latent is temporally concatenated with the noisy video latents, with its corresponding mask set to 1 to explicitly designate it as the visual context.

\noindent \textbf{Motion-Shape Disentanglement.} During inference, the reference image and the driving video often originate from distinct subjects. Directly applying the tracked mesh of the driving video would inevitably leak the driving subject's body shape into the generated video (e.g., imposing a robust physique onto a slender reference). To address this, we implement an explicit motion-shape disentanglement strategy. Specifically, we retarget the driving mesh by retaining the pose and expression from the driving video, while injecting the shape parameters extracted from the reference image. The retargeted mesh $M_{ehm}^{target}$ is formalized as:
\begin{equation}
M_{ehm}^{target} = M_{ehm}(\theta^{d}, \psi^{d}, \beta_b^{r}, \beta_f^{r})~,
\label{eq:ehm_rebuild}
\end{equation}
where $\theta^{d}$ and $\psi^{d}$ denote the pose and expression parameters from the driving video, respectively, while $\beta_b^{r}$ and $\beta_f^{r}$ represent the body and facial shape parameters estimated from the reference image. Additionally, the camera parameter $C^d$ from the driving video is applied during rendering. Through this parameter recombination, we ensure the generated motion remains faithful to the driving video, while the physical shape is strictly anchored to the reference subject.

\subsection{Inference and Training}
\label{subsec:infer_train}
\noindent \textbf{Long Video Inference.} Due to GPU memory constraints, generating long videos in a single pass is intractable. We thus adopt an autoregressive strategy based on overlapping frames. Specifically, the last five frames of the preceding chunk are encoded into temporal latents and serve as the temporal context for the subsequent chunk. These latents are concatenated after the reference latent with their mask values set to 1. To ensure character consistency across the entire video, following~\cite{flashtalk,histream}, we fix the original reference latent as a "visual anchor" at the beginning of the latent sequence for every generated chunk.

\noindent \textbf{Spatially-Aligned Conditioning.} We identify the spatial misalignment between training and inference as a critical factor contributing to identity degradation in long video generation. During training, the reference image and target frames originate from the same video, exhibiting high spatial alignment. Conversely, cross-identity inference introduces spatial discrepancies between the reference image and driving motion. This training-inference gap biases the model to attend more to the spatially aligned temporal latents, gradually neglecting the information in the reference latent. Consequently, as autoregressive errors accumulate, the generated video suffers from visual artifacts and identity drift.

To mitigate this issue, we propose a Spatially-Aligned Conditioning strategy, as illustrated in \cref{fig:SAC}. After generating the first video chunk, we retain the latent of its initial frame. For all subsequent chunks, this spatially-aligned latent is inserted between the original reference latent and the temporal latents, with its mask set to 1. Acting as a structural bridge, it preserves both the original identity and the spatial layout of the target scene. Furthermore, since this latent accumulates fewer errors than the recursively generated temporal latents, it acts as a reliable visual anchor to redistribute the model's attention. This prevents over-reliance on noisy temporal cues, thereby significantly enhancing appearance consistency and reducing artifacts in long video generation.

\noindent \textbf{Training.} To adapt the model for both cold-start and autoregressive generation, we probabilistically toggle the use of temporal latents. Additionally, to simulate spatial misalignment, we randomly sample reference frames from the video, apply simple geometric augmentations (e.g., rotation, translation), and randomly inject an auxiliary reference latent, mirroring the Spatially-Aligned Conditioning setup used during inference. Finally, we train our model using the standard Flow Matching~\cite{flow_matching} objective with an Optimal Transport path.

\section{Experiment}
\label{subsec:exper}
\subsection{Setup}
\noindent \textbf{Implementation Details.} Our model is implemented in PyTorch~\cite{pytorch} and trained using Adam~\cite{adam}. Training is conducted on 32 NVIDIA H20 GPUs with a global batch size of 32. The entire training process spans approximately 100 hours, covering roughly 6,500 iterations to reach convergence. During training, we randomly sample 77-frame video clips at a resolution of $512 \times 512$, paired with a reference image as the conditioning input. Our backbone adopts a DiT~\cite{dit} architecture consistent with Wan2.1-I2V~\cite{wan}. Since Wan-Animate~\cite{wan-animate} shares this structure and has been pre-trained on large-scale human videos, we initialize our DiT module with its weights to leverage this strong prior. For the Motion Tracker, we utilize GVHMR~\cite{gvhmr}, TEASER~\cite{teaser}, and HaMeR~\cite{hamer} for the coarse estimation of body pose, facial expressions, and hand parameters, respectively. DWPose~\cite{dwpose} and MediaPipe~\cite{mediapipe} are employed to extract body and facial keypoints for tracking refinement.



\noindent \textbf{Training Dataset.} We construct a composite dataset comprising approximately 900k video clips. The core data originates from human motion videos (37.8\%) crawled from the Internet and the Speaker-Vid~\cite{speakervid} dataset (60.6\%). To ensure high visual quality and motion dynamics, we rigorously filter these sources by discarding clips with undetected faces, insufficient face scales, low hand visibility, or static subjects. Finally, to enhance facial animation, we incorporate the VFHQ~\cite{vfhq} dataset (1.6\%) into our final training corpus.

\noindent \textbf{Baselines.}
We compare EMOSH against several state-of-the-art (SOTA) human animation methods, including Wan-Animate\cite{wan-animate}, HyperMotion \cite{hypermotion}, MimicMotion~\cite{mimicmotion}, Moore-AA (AnimateAnyone by Moore)~\cite{moore2024animateanyone}, StableAnimator~\cite{stableanimator}, UniAnimate-DiT and UniAnimate~\cite{unianimate}, and Champ~\cite{champ}. To ensure a fair comparison, we conduct inference for all baseline models at their native resolutions and subsequently evaluate metrics on a unified spatial scale.

\noindent \textbf{Evaluation Protocols. }
We evaluate our model under two distinct settings: Self-driven and Cross-driven.
\textit{Self-driven:} The driving video and the reference image originate from the same video. We utilize the public benchmarks EchoMimicV2~\cite{echomimicv2} and TikTok~\cite{tiktokdataset} datasets, alongside a self-collected test set containing 150 videos ($\sim$55k frames). For evaluation, the first frame of the video serves as the reference image, and the model reconstructs the full video sequence based on the driving signals. To quantitatively evaluate the reconstruction quality, we employ PSNR, L1, SSIM, and LPIPS~\cite{lpips} to compare the generated videos against the ground truth.
\textit{Cross-driven:} We collect 12 reference images covering diverse body shapes (e.g., varying heights and weights) and 10 driving videos featuring distinct motions. These are cross-paired to generate a total of 120 video samples ($\sim$51k frames). Due to the absence of Ground Truth, we primarily rely on Human Evaluation. Additionally, we use the Identity Preservation Score (IPS) as a supplementary metric. Specifically, we utilize ArcFace~\cite{arcface} to extract facial features from both the generated video and the reference image, calculating the average cosine similarity between them.

\begin{table}[tb]
\caption{Quantitative results on three datasets under the self-driven setting. Our method achieves the best performance across all evaluation metrics.}
\label{tab:self_driven}
\centering
\resizebox{\linewidth}{!}{
\begin{tabular}{c cccc cccc cccc c}
    \toprule
    \multirow{2}{*}{Method} & \multicolumn{4}{c}{EchoMimicV2 dataset} & \multicolumn{4}{c}{TikTok dataset} & \multicolumn{4}{c}{Self-collected dataset} & \multirow{2}{*}{condition} \\
    \cmidrule(lr){2-5} \cmidrule(lr){6-9} \cmidrule(lr){10-13}
    & PSNR$\uparrow$ & L1$\downarrow$ & SSIM$\uparrow$ & LPIPS$\downarrow$ & PSNR$\uparrow$ & L1$\downarrow$ & SSIM$\uparrow$ & LPIPS$\downarrow$ & PSNR$\uparrow$ & L1$\downarrow$ & SSIM$\uparrow$ & LPIPS$\downarrow$ \\
    \midrule
    Moore-AA & 20.39 & 0.0513 & 0.7388 & 0.2059 & 16.23 & 0.1022 & 0.6742 & 0.3263 & 18.11 & 0.0762 & 0.6464 & 0.2788 & 2D Pose\\
    StableAnimator & 16.76 & 0.0796 & 0.6724 & 0.2836 & 11.76 & 0.1845 & 0.5732 & 0.4437 & 15.97 & 0.1024 & 0.6179 & 0.3285  & 2D Pose\\
    MimicMotion & 19.52 & 0.0622 & 0.7211 & 0.2450 & 14.17 & 0.1335 & 0.3876 & 0.6316 & 16.00 & 0.1057 & 0.6179 & 0.3552 & 2D Pose\\
    UniAnimate & 16.39 & 0.1045 & 0.6933 & 0.2678 & 11.73 & 0.1939 & 0.6096 & 0.4320 & 14.69 & 0.1286 & 0.6048 & 0.3543 & 2D Pose\\
    UniAnimate-DiT & 17.90 & 0.0678 & 0.7266 & 0.2533 & 12.79 & 0.1571 & 0.6151 & 0.4213 & 14.76 & 0.1302 & 0.5778 & 0.3688 & 2D Pose\\
    HyperMotion & 22.32 & 0.0391 & 0.7975 & 0.1800 & 15.48 & 0.1154 & 0.6529 & 0.3536 & 19.48 & 0.0660 & 0.7082 & 0.2512 & 2D Pose\\
    Wan-Animate & 22.86 & 0.0379 & 0.7750 & 0.1802 & 17.18 & 0.0932 & 0.6808 & 0.3209 & 20.72 & 0.0555 & 0.7281 & 0.2229 & 2D Pose \& Face\\
    \midrule
    Champ & 17.40 & 0.0756 & 0.6502 & 0.2911 & 13.34 & 0.1482 & 0.5902 & 0.4156 & 14.85 & 0.1159 & 0.5602 & 0.3727 & Mesh \\
    EMOSH (Ours) & \bf 23.66 & \bf 0.0308 & \bf 0.8240 & \bf 0.1428  & \bf 17.80 & \bf 0.0831 & \bf 0.740 & \bf 0.2933  & \bf 21.18 & \bf 0.0522 & \bf 0.751 & \bf 0.2066 & Mesh  \\
    \bottomrule
\end{tabular}
}
\end{table}

\begin{table}[tb]
\caption{Comparison of Identity Preservation Score (IPS) in the cross-driven setting. Our method better maintains ID consistency.}
\resizebox{1.0\linewidth}{!}{
\begin{tabular}{cccccccccc}
    \toprule
       & Moore-AA & Champ & StableAnimator & MimicMotion & UniAnimate & UniAnimate-DiT & HyperMotion & Wan-Animate & EMOSH (Ours)   \\
    \midrule
    IPS$\uparrow$  & 0.2174 & 0.2619 & 0.0935 & 0.0744 & 0.1618 & 0.3130 & 0.2872 & 0.3802 & \bf 0.4445  \\
   \bottomrule
\end{tabular}
}
\label{tab:cross_driven_ips}
\end{table}

\begin{figure}[tb]
  \centering
  \includegraphics[width=\textwidth]{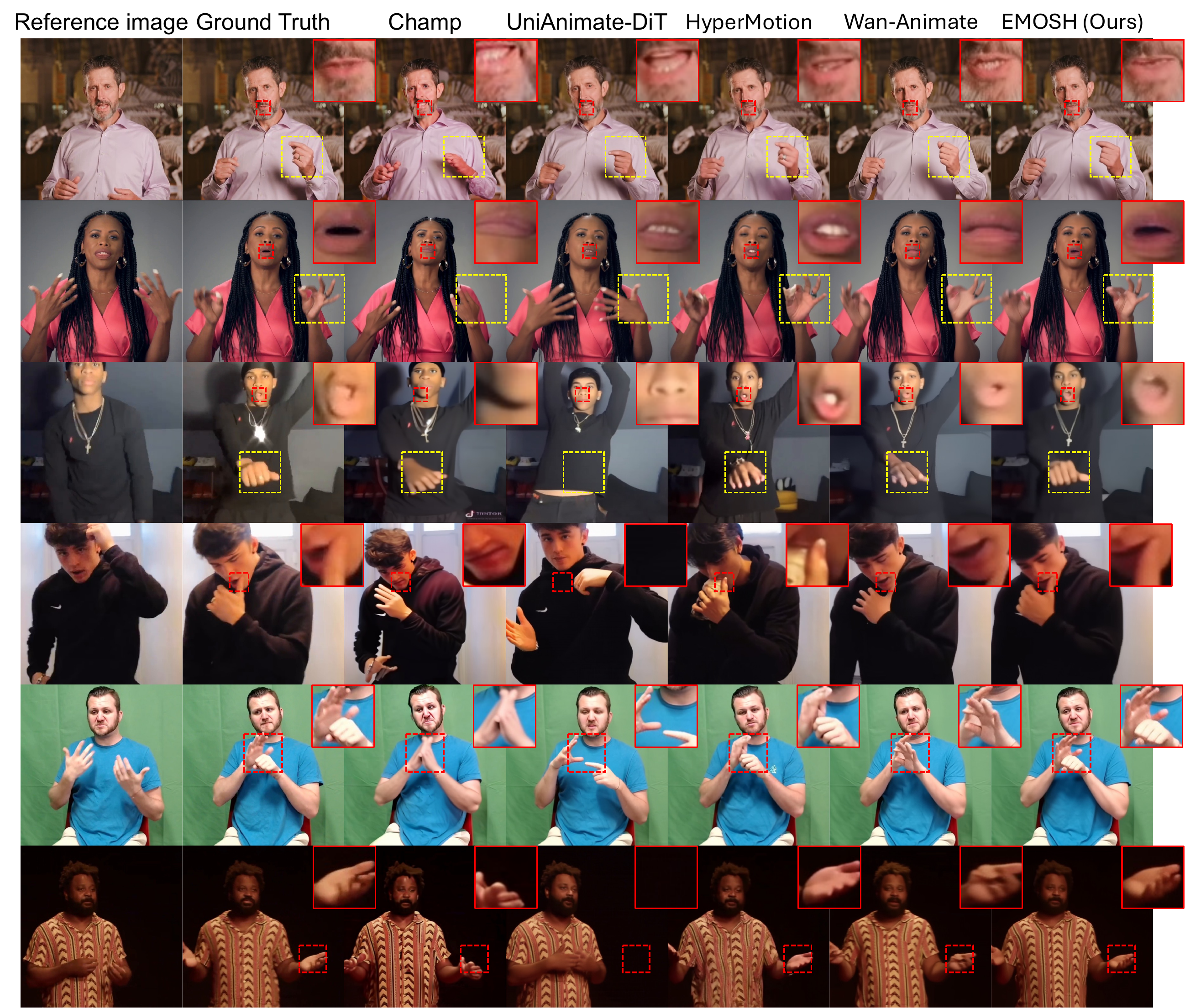}
  \caption{
  Qualitative results on three datasets under the self-driven setting. EMOSH generates higher-fidelity videos and achieves more accurate control of facial expressions and hand gestures compared to the baselines.
  }
  \label{fig:compare_selfact}
\end{figure}

\begin{figure}[tb]
  \centering
  \includegraphics[width=\textwidth]{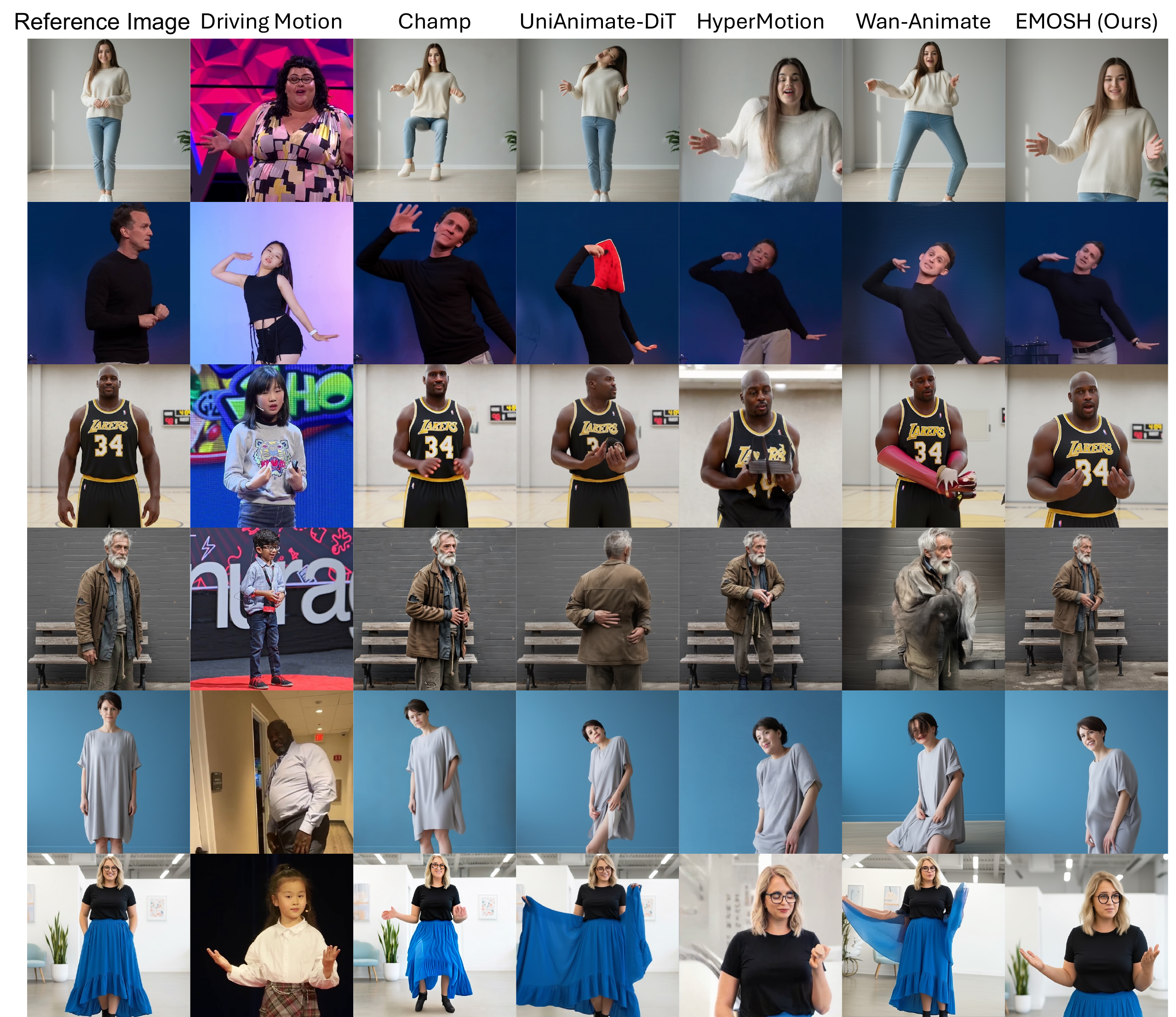}
  \caption{
Visual quality comparison under the cross-driven setting. While achieving precise motion control, our method better preserves the identity and body shape characteristics of the reference subject, effectively avoiding the shape distortion and limb artifacts seen in other methods.
  }
  \label{fig:compare_crossact}
\end{figure}

\begin{table*}[t]
\caption{Human evaluation results are based on the standard GSB (Good/Same/Bad) protocol for pairwise comparisons between our approach and three baseline methods. The values and colored bars indicate the percentage of user preferences for \textcolor{OursWin}{\textbf{Ours}}, \textcolor{BothSame}{\textbf{Both (Same)}}, and \textcolor{OursLoss}{\textbf{Competitor}}.}
  \centering
  \resizebox{\textwidth}{!}{
  \begin{tabular}{@{} c @{\hspace{3em}} c @{\hspace{3em}} c @{}}
    \toprule
    \textbf{Ours vs Wan-Animate} & \textbf{Ours vs HyperMotion} & \textbf{Ours vs UniAnimate-DiT} \\
    \midrule
    \TriBar{0.8921}{0.0396}{0.0683} & \TriBar{0.9842}{0.0075}{0.0083} & \TriBar{0.9562}{0.0180}{0.0258} \\
    \addlinespace[3pt]
    \textcolor{OursWin}{\textbf{89.21\%}} \quad \textcolor{BothSame}{\textbf{3.96\%}} \quad \textcolor{OursLoss}{\textbf{6.83\%}} &
    \textcolor{OursWin}{\textbf{98.42\%}} \quad \textcolor{BothSame}{\textbf{0.75\%}} \quad \textcolor{OursLoss}{\textbf{0.83\%}} &
    \textcolor{OursWin}{\textbf{95.62\%}} \quad \textcolor{BothSame}{\textbf{1.80\%}} \quad \textcolor{OursLoss}{\textbf{2.58\%}} \\
    \bottomrule
  \end{tabular}
  }
  \label{tab:human_eval}
\end{table*}

\subsection{Quantitative Results}
\label{subsec:quantitative_results}
\noindent \textbf{Self-driven.}
\cref{tab:self_driven} summarizes the quantitative comparisons between our method and the baseline models across three datasets. Our method achieves the best performance across all metrics. This demonstrates that our approach can generate higher-fidelity videos conditioned on the driving video and reference image while maintaining precise motion control.

\noindent \textbf{Cross-driven.}
\cref{tab:cross_driven_ips} presents the Identity Preservation Score (IPS) under the Cross-ID setting. The results show that our approach achieves the highest IPS, verifying its superior robustness in preserving the identity features of the reference image under diverse driving poses.

\subsection{Qualitative Results}
\label{subsec:qualitative_result}
\noindent \textbf{Self-driven.}
As illustrated in \cref{fig:compare_selfact}, we present the qualitative comparison of various methods across three datasets under the self-driven setting. While most approaches can reasonably preserve the identity of the reference image, they struggle with fine-grained details. Specifically, Champ's reliance on a human mesh control mechanism makes it difficult to achieve fine-grained control over expressions and gestures. UniAnimate-DiT, although utilizing 2D poses, lacks facial keypoints, leading to inaccurate expression control. Moreover, it may struggle with complex body poses and gestures, resulting in failure on some instances. HyperMotion and Wan-Animate produce videos of higher overall quality and can generally control expressions and gestures. However, they still suffer from incorrect mouth shapes, and blurry or merged finger artifacts. In contrast, although our method is mesh-guided, it achieves control performance that is comparable to, or even surpasses, methods relying on fine-grained 2D poses. This demonstrates our method generates videos with superior accuracy and higher fidelity.

\noindent \textbf{Cross-driven.}
As shown in \cref{fig:compare_crossact}, we present qualitative results under the cross- driven setting. Some baselines struggle to preserve the reference identity, causing shape distortions like unnatural thinning or widening. Some also fail to follow the driving video's camera framing, rigidly keeping the original body scale.
Specifically, Champ fails to accurately control expressions and gestures. UniAnimate-DiT struggles with expression driving and frequently generates merged finger artifacts. HyperMotion improves expression control but lacks ID preservation, leading to appearance deformations. Although Wan-Animate uses 2D pose retargeting to mitigate shape leakage, it still generates disproportionate bodies, or even abnormal limb lengths with severe artifacts in some cases.
In contrast, our EMOSH demonstrates remarkable superiority. It perfectly disentangles and preserves the reference's authentic body shape and identity while achieving highly accurate control over poses, expressions, and camera framing.

\subsection{Human Evaluation}
To further subjectively assess visual performance in the cross-driven setting, we conducted a user study comparing EMOSH against recent DiT-based state-of-the-art open-source baselines: Wan-Animate, HyperMotion, and UniAnimate-DiT. We designed a pairwise comparison test with 20 participants. In each trial, participants viewed a reference image, a driving video, and two anonymized, randomly ordered generated videos. They were instructed to select their preferred result based on three criteria: video generation quality, motion and expression accuracy, and identity preservation. As reported in \cref{tab:human_eval}, EMOSH consistently obtains the highest user preference, further validating that our approach produces superior video results that better align with human visual perception.

\begin{table}[tb]
  \centering
  \caption{Quantitative results of our ablation study. (Left) Under the self-driven setting, removing the Tracker or Hybrid Motion leads to a significant drop in generation metrics. (Right) Under the cross-driven setting, removing Disentanglement or Spatially-Aligned Conditioning (SAC) degrades the identity preservation score.}
  \label{tab:overall_ablation}
  \begin{subtable}[c]{0.67\linewidth}
    \centering
    \label{tab:my_left_table}
    \resizebox{\linewidth}{!}{
    \begin{tabular}{c cccc cccc}
      \toprule
      \multirow{2}{*}{Method} & \multicolumn{4}{c}{EchoMimicV2 dataset} & \multicolumn{4}{c}{Self-collected dataset}  \\
      \cmidrule(lr){2-5} \cmidrule(lr){6-9} 
      & PSNR$\uparrow$ & L1$\downarrow$ & SSIM$\uparrow$ & LPIPS$\downarrow$ & PSNR$\uparrow$ & L1$\downarrow$ & SSIM$\uparrow$ & LPIPS$\downarrow$  \\
      \midrule
      w/o Tracker       & 20.51 & 0.0434 & 0.7789 & 0.1909 & 16.48 & 0.0959 & 0.7095 & 0.3300 \\
      w/o Hybrid Motion & 22.90 & 0.0345 & 0.8127 & 0.1492 & 17.55 & 0.0850 & 0.7330 & 0.2973 \\
      Full (Ours)       & \bf 23.66 & \bf 0.0308 & \bf 0.8240 & \bf 0.1428  & \bf 17.80 & \bf 0.0831 & \bf 0.740 & \bf 0.2933  \\
      \bottomrule
    \end{tabular}
    }
  \end{subtable}\hfill
  \begin{subtable}[c]{0.31\linewidth}
    \centering
    \label{tab:my_right_table}
    \resizebox{\linewidth}{!}{
    \begin{tabular}{lc}
      \toprule
      Method & IPS$\uparrow$   \\
      \midrule
      w/o Disentanglement & 0.4258  \\
      w/o SAC             & 0.4237 \\
      Full (Ours)         & \bf 0.4445 \\
      \bottomrule
    \end{tabular}
    }
  \end{subtable}
\end{table}

\begin{figure}[tb]
  \centering
  \begin{subfigure}[b]{0.44\textwidth}
    \centering
    \includegraphics[width=\textwidth]{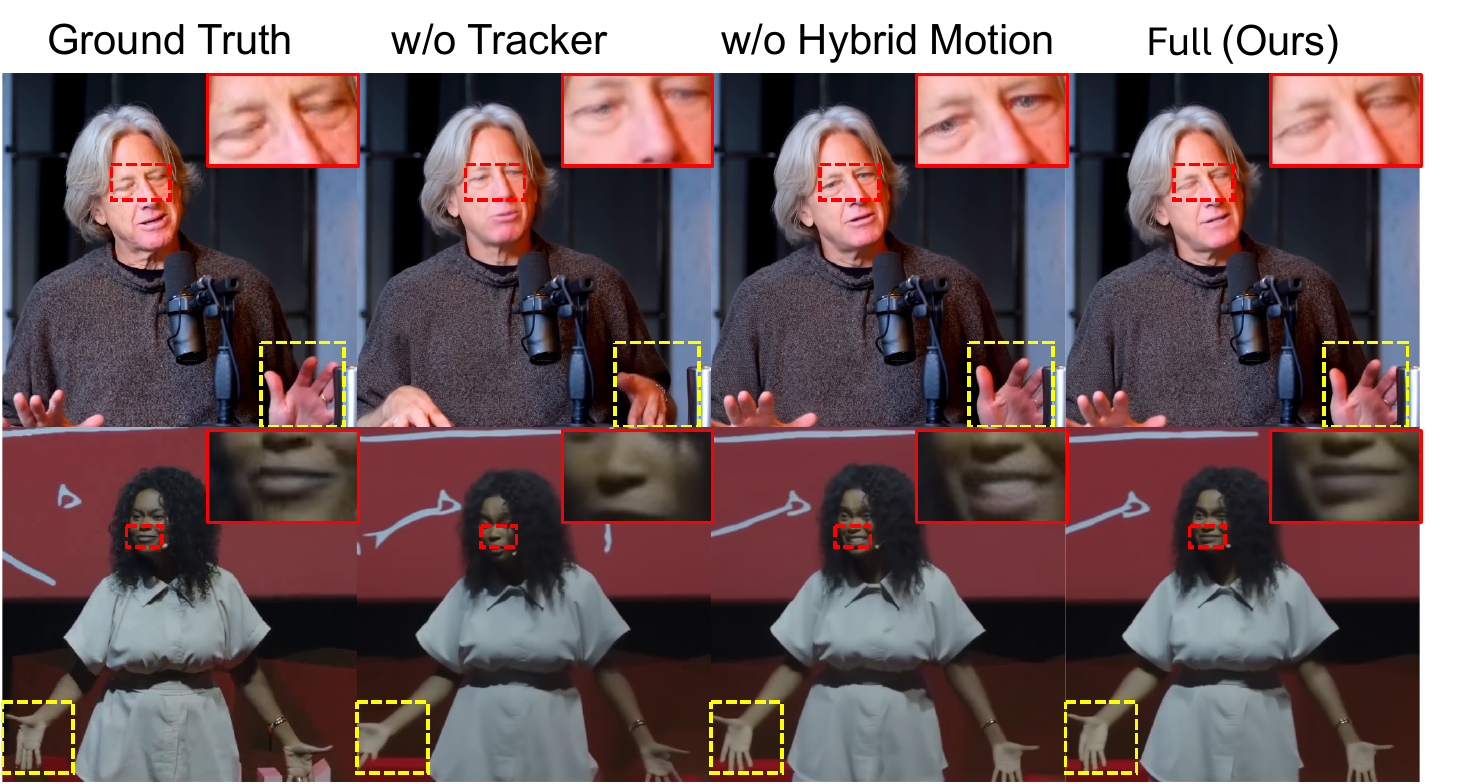}
  \end{subfigure}
  \hfill
  \begin{subfigure}[b]{0.55\textwidth}
    \centering
    \includegraphics[width=\textwidth]{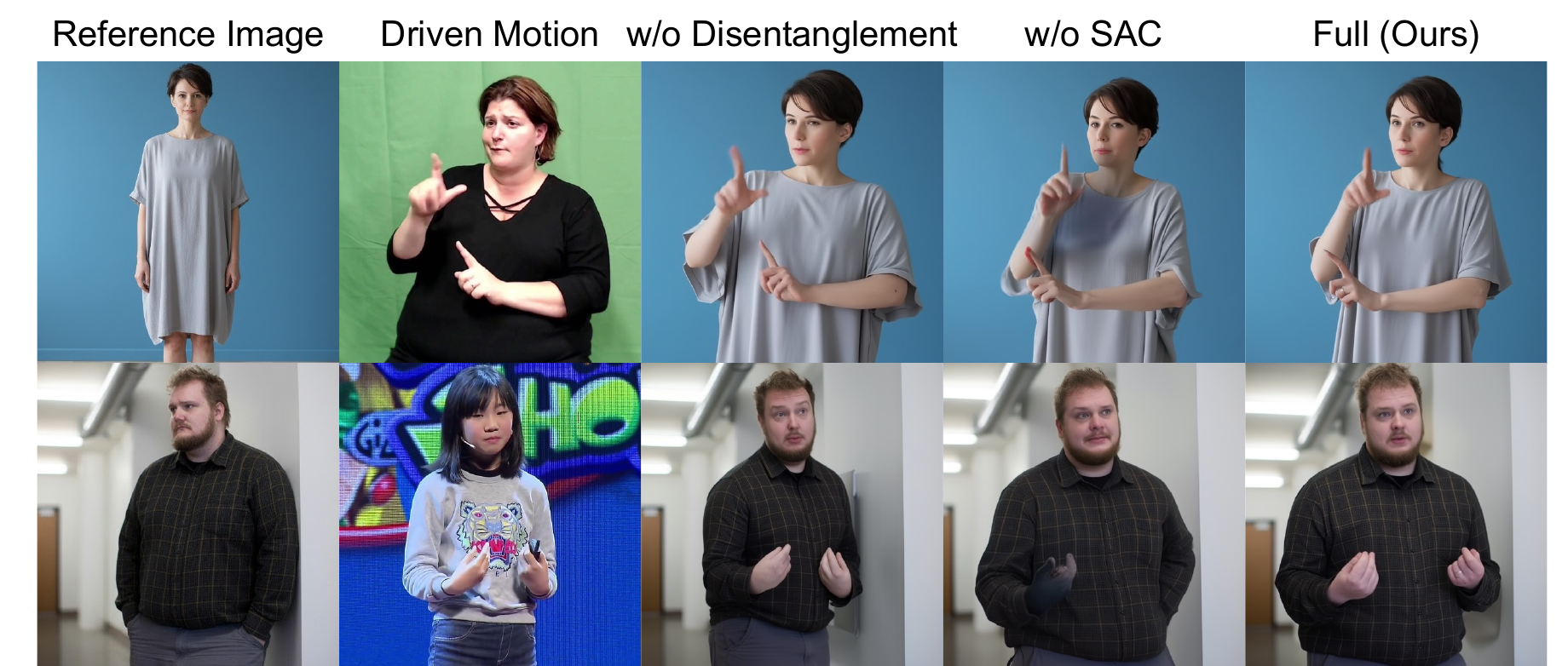}
  \end{subfigure}
  \caption{Qualitative ablation results. (Left) Under the self-driven setting, full method enables more accurate control over expressions and gestures. (Right) Under the cross-driven setting, full method disentangles shape and motion, whereas removing Spatially-Aligned Conditioning (SAC) leads to artifacts and reduces identity preservation.}
  \label{fig:ablation}
\end{figure}

\subsection{Ablation Studies}
To verify the contribution of each core component, we perform ablation experiments under the self-driven and cross-driven settings, with quantitative and qualitative results detailed in \cref{tab:overall_ablation} and \cref{fig:ablation}.

\noindent \textbf{w/o Tracker.}
By removing the proposed Motion Tracker, this baseline relies on off-the-shelf models for motion estimation, akin to the strategy used in Champ. As illustrated in \cref{fig:ablation}, it struggles to accurately control facial expressions and intricate hand gestures. Furthermore, it suffers a significant performance drop across all metrics in the self-driven setting, as shown in \cref{tab:overall_ablation}.

\noindent \textbf{w/o Hybrid Motion.}
This baseline discards 2D keypoints drawing operation and depends on naive mesh renderings for motion conditioning. Visually, it struggles to animate facial expressions like blinking, and its gesture control granularity falls short of the full method. Additionally, its quantitative metrics under the self-driven setting drop significantly.

\noindent \textbf{w/o Disentanglement.}
By removing the EHM Retargeting, this baseline discards the motion-shape disentanglement. Without this, the model suffers from shape leakage and fails to preserve the reference body shape. This directly translates to a noticeable drop in identity consistency, reflected by the lower IPS score. 

\noindent \textbf{w/o SAC.}
This variant removes the Spatially-Aligned Conditioning (SAC) mechanism. Consequently, the model produces artifacts during video generation. Furthermore, it weakens the model's capability to preserve identity features, as evidenced by the degraded IPS metric.

\section{Conclusion}
In this paper, we propose EMOSH, a high-fidelity controllable human video generation framework that achieves expressive motion and shape disentanglement. By introducing the Expressive Human Model as the core motion control representation, we explicitly disentangle the motion from the body shape in the driving signal. Coupled with our proposed Confidence-Aware Motion Tracker, EMOSH enables fine-grained control over human poses, facial expressions, and complex hand gestures in video generation. This successfully overcomes the inherent bottleneck of traditional mesh-guided generation methods, which typically struggle to achieve precise control over facial and hand movements. To further enhance control precision, we propose a Coarse-to-Fine Hybrid Motion Injection strategy. Finally, our Spatially-Aligned Conditioning mechanism effectively bridges the domain gap between training and inference, significantly reducing visual artifacts and enhancing identity preservation in cross-driven scenarios. Extensive experiments demonstrate the superiority of EMOSH in control precision, identity preservation, and overall visual quality. Further discussions about the limitations of our method are provided in the supplementary material.

\bibliographystyle{splncs04}
\bibliography{main}

\author{}
\institute{}
\title{EMOSH: Expressive Motion and Shape Disentanglement for Human Animation} 
\subtitle{Supplementary Material}
\titlerunning{EMOSH}
\maketitle
\thispagestyle{empty}
\appendix

\section*{Overview}
\noindent This supplementary material presents more details and additional results not included in the main paper due to page limitation. The list of items included is:
\begin{itemize}
\item Video demo at \href{https://eastbeanzhang.github.io/EMOSH/}{\textcolor{magenta}{Project page}} with a brief description in \cref{spsec:videodemo}.
\item More model implementation details in \cref{spsec:more-im-de}.
\item More tracker details in \cref{spsec:tracking}.
\item More experimental results in \cref{spsec:more-results}.
\item Further discussion of limitation and ethical considerations in \cref{spsec:more-discuss}.
\end{itemize}

\section{Video Demo}
\label{spsec:videodemo}
We highly encourage readers to view the video demo included in our \href{https://eastbeanzhang.github.io/EMOSH/}{\textcolor{magenta}{Project page}}. The video showcases EMOSH's capability in generating expressive human animations conditioned on the 3D EHM~\cite{guava}, as well as basic zoom-in and zoom-out camera controls. Additionally, we provide visual comparisons between EMOSH and state-of-the-art open-source baselines, including UniAnimate-DiT~\cite{unianimate}, HyperMotion~\cite{hypermotion}, and Wan-Animate~\cite{wan-animate}, under cross-driven setting. We also briefly present visual results from our ablation studies and human tracking comparisons. The video demonstrates that our 3D Mesh-guided approach achieves the same level of expressive motion control as 2D pose-based methods, while successfully ensuring shape disentanglement and yielding superior visual quality.

\section{More Implementation Details}
\label{spsec:more-im-de}
In this section, we provide further details regarding our model architecture and training procedures.

\subsection{More Preliminaries}
We utilize the Flow Matching~\cite{flow_matching} paradigm for training. This simulation-free continuous normalizing flow method demonstrates superior efficiency and stability in video generation tasks. It aims to learn a time-dependent velocity field $v(\cdot)$, defining the mapping trajectory from the noise distribution $p_1$ to the data distribution $p_0$. The generation process is described by an Ordinary Differential Equation (ODE):
\begin{equation}
\frac{dz_t}{dt}=v(z_t,t,c),
\end{equation}
where $c$ represents conditional information (e.g., image, text) and $z_t$ denotes the latent state at timestep $t$. During training, we adopt the Optimal Transport (OT) path (i.e., linear path) as the probability path $p_t(z)$. By linearly interpolating between data latent $z_0$ and noise $z_1 \sim \mathcal{N}(0,I)$, we obtain the intermediate state:
\begin{equation}
z_t=(1-t)\cdot z_0+t\cdot z_1,
\end{equation}
Along this path, the ground-truth velocity for any state is $z_1 - z_0$. Consequently, the training objective is to regress the model-predicted velocity field to this ground truth:
\begin{equation}
L_{FM}=\mathbb{E}_{t,z_0,z_1}[\Vert v(z_t,t,c)-(z_1-z_0)\Vert ^2 ]
\end{equation}
During inference, starting from randomly sampled Gaussian noise $z_1$, generation is achieved by solving the aforementioned ODE from $t=1$ to $t=0$ using a numerical solver.

\subsection{Model details}
We build our framework based on the Wan2.1-I2V-14B~\cite{wan} model. To ensure efficient training, we opt not to fine-tune all the parameters. Instead, we focus on updating the Diffusion Transformer (DiT) module~\cite{dit} using Low-Rank Adaptation (LoRA)~\cite{lora}. Specifically, we set the LoRA rank to 32 and the learning rate to $1 \times 10^{-4}$.
Furthermore, as the 16-dimensional motion features extracted by the VAE cannot be directly injected into the DiT, we train a dedicated motion embedding layer to bridge this gap. This layer projects the 16-dimensional VAE features into a 5120-dimensional space to align with the input requirements of the DiT.

\subsection{Training Details}
To adapt the model for different generation scenarios, we apply temporal latents 50\% of the iterations during training. For the remaining 50\%, the model is trained using only the reference latent. To improve the model's robustness, we apply simple geometric augmentations to the reference image with a 50\% probability, which includes slight random rotations, translations, and cropping.
Furthermore, with temporal reference latents enabled, we sample an extra video frame with a 10\% probability to serve as an auxiliary reference. This latent is interposed between the original reference and the temporal latents, allowing the model to accommodate the spatially aligned latent used during inference. Finally, we exclude the reference latents from the loss computation. Correspondingly, the motion information is injected solely into the latents that follow the reference latent.

\subsection{Evaluation Details}
The baseline models used for comparison generate videos at various aspect ratios, including portrait, square, and landscape formats. To ensure a fair and standardized comparison, we uniformly evaluate all methods at our adopted square resolution. For video generation, we run all baseline models using their default parameter configurations. Specifically, we first pad the input reference images and driving videos to match the native resolution required by each respective baseline. Following inference, we crop the generated videos back to the unified square resolution before computing the evaluation metrics.

\subsection{Motion-Shape Disentanglement}
As detailed in the main paper \cref{subsec:motion_control}, we achieve disentanglement by replacing the shape parameters of the driving video's tracked EHM with the reference image's shape parameters $\beta^r$. However, a direct replacement inevitably alters the overall body length. Consequently, when rendering the condition map using the driving video's original camera parameters $C$, significant differences in body proportions can cause the head position to shift—sometimes even pushing the face partially off-screen.
To resolve this issue, we calculate the difference in the average y-coordinate of the head between the first frame of the original driving mesh and that of the shape-replaced mesh. By applying this difference as a global y-axis offset to the entire shape-replaced mesh, we effectively eliminate facial misalignment and ensure the subject remains properly framed.

\section{Motion Tracking}
\label{spsec:tracking}
In the main paper \cref{subsec:motion_tracking}, we briefly introduced our confidence-aware motion tracker, designed to accurately extract human motions, including body poses, facial expressions, and complex hand gestures. Here, we provide detailed implementation specifics for this tracking pipeline. Our tracker operates in a two-stage paradigm. Coarse Initialization and Confidence-Aware Joint Optimization. We adopt the Expressive Human Model (EHM)~\cite{guava} as our foundational 3D parametric representation, which unifies SMPL-X~\cite{smplx} and FLAME~\cite{flame}.

\subsection{Coarse Initialization.}
Given a monocular driving video, we first perform feature extraction to estimate the initial pose parameters and 2D guidance. This serves as a reliable prior for the subsequent joint optimization process.

\noindent \textbf{Dense 2D Landmarks Extraction.}
We utilize DWPose~\cite{dwpose} to extract human keypoints. To compensate for the limitations of sparse DWPose keypoints in capturing facial expressions, we extract additional dense 2D facial keypoints. Specifically, we obtain 478 MediaPipe~\cite{mediapipe} facial landmarks to provide robust 2D supervision signals for eye gaze and micro-expressions.

\noindent \textbf{Bounding Box and Cropping.}
We compute a full-body bounding box based on the detected human keypoints to isolate the central human region. To capture fine-grained local details, we extract additional crops for the head and hands using their respective 2D facial and hand keypoints.

\noindent \textbf{Human Pose Estimation.}
We leverage off-the-shelf estimators to extract coarse 3D geometric priors. Specifically, we run GVHMR~\cite{gvhmr} on the full-body crops to obtain initial body pose and projection parameters, which are then converted into perspective camera parameters. For local details, we apply HaMeR~\cite{hamer} on the hand crops to predict MANO parameters, and utilize TEASER~\cite{teaser} on the head crops to obtain initial FLAME expression and jaw poses parameters.

\subsection{Confidence-Aware Joint Optimization}
Since the features derived from the three independent feed-forward estimators cannot be directly integrated and exhibit spatial misalignment, we propose an optimization-based joint framework. This framework optimizes the full set of EHM and camera parameters, denoted as $\Theta = \{ \theta_{body}, \theta_{hand}, \theta_{jaw}, \psi, \beta\}$ and $C$, guided by the estimated 2D keypoints and human pose to accurately track motions. The overall objective function is formulated as follows:
\begin{equation}
    \mathcal{L}_{total} = \mathcal{L}_{2D} + \mathcal{L}_{3D\_hand} + \mathcal{L}_{reg} + \mathcal{L}_{smooth}.
\end{equation}

\noindent During the 2D-to-3D fitting process, challenges such as jitter, self-occlusion, and extreme poses often lead to severe errors in 2D keypoint estimation. To mitigate these issues, we incorporate a confidence-aware dynamic weighting strategy into $\mathcal{L}_{2D}$. This strategy leverages both the confidence scores from the 2D detector and explicit verifications derived from 3D geometric priors.

\noindent \textbf{Confidence Score Weighting.}
We first filter the keypoints based on the detection scores provided by DWPose. To prevent the model from fitting to erroneous noise, the loss weights for keypoints with confidence scores below a predefined threshold are set to 0. Meanwhile, to maintain torso stability, we amplify the fitting weights for core joints (e.g., shoulders and pelvis). For limb keypoints, the weights are also increased if their confidence exceeds the threshold.

\noindent \textbf{Explicit Face Visibility Check.}
 When dealing with large head rotations (e.g., profile or back views), 2D facial keypoints are highly susceptible to feature drift. To mitigate this, we introduce a dynamic visibility check. Specifically, we compute and normalize the face direction vector and calculate the angle relative to the camera's z-axis. We then evaluate its three spatial components: yaw, pitch, and roll. If any of these angles exceeds a predefined threshold, the face is classified as severely rotated or invisible. Consequently, all face-related loss weights are masked to zero. This effectively prevents facial distortion caused by the optimizer's attempt to fit the mesh to erroneous facial keypoints.

\noindent \textbf{Dynamic Hand Validity Check.}
Hands exhibit the highest degrees of freedom in human motion and are frequently subject to occlusion. Relying solely on confidence scores for filtering often fails when hands exit the frame but erroneously reappear during optimization. To address this, we propose a joint validation mechanism based on 3D-to-2D projected bounding boxes. During each optimization iteration, we project the current 3D hand keypoints onto the 2D image plane to derive a reference bounding box. We then compute the Intersection over Union (IoU) between this projected box and the image boundaries. If the IoU falls below a predefined threshold or a valid match cannot be established, the hand state is flagged as "Invalid." For these cases, we down-weight the 2D keypoint loss and replace the coordinates with projections from smoothed 3D hand priors, effectively shifting them out of the frame. This ensures that during severe occlusions, hand tracking falls back to robust 3D priors rather than relying on erroneous 2D guidance.

\noindent \textbf{2D Reprojection Loss ($\mathcal{L}_{2D}$).}
To ensure strict alignment between the 3D model and the image, we project the 3D EHM keypoints $J_{3D}$ onto the 2D image plane via perspective projection $\Pi_{K, R, t}(\cdot)$. The loss against the detected 2D keypoints $J_{2D}$ is then calculated as follows:
\begin{equation}
\mathcal{L}_{2D} = \sum_{k \in \mathcal{K}} w_k \| \Pi_{K, R, t}(J_{3D}^k) - J_{2D}^k \|_1
\end{equation}
where $\mathcal{K}$ denotes the set of keypoints, encompassing the DWPose body joints, dense facial landmarks, and hand keypoints. Notably, the weight $w_k$ is not static. Instead, it is dynamically determined by our previously described confidence-aware weighting strategy and explicit face or hand validity checks. This design empowers the model to disregard unreliable 2D gradient guidance when encountering occlusions or extreme viewpoints.

\noindent \textbf{3D Hand Vertex Alignment Loss ($\mathcal{L}_{3D\_hand}$).}
In complex hand gestures (e.g., making a fist or crossing fingers), relying solely on 2D keypoints introduces severe depth ambiguity. To address this, we introduce an explicit 3D vertex alignment loss. Specifically, we constrain the optimized EHM hand vertices $V_{hand}$ to align with the reference vertices $V_{ref}$ reconstructed from the HaMeR prior:
\begin{equation}
\mathcal{L}_{3D\_hand} = \sum_{v \in \mathcal{V}_{hand}} \mathbb{I}_{hand} \cdot \left( \| V_{hand}^v - V_{ref}^v \|_1 + \lambda_{z} \| V_{hand, z}^v - V_{ref, z}^v \|_1 \right)
\end{equation}
Here, $\mathbb{I}_{hand}$ is an indicator function derived from the aforementioned "Dynamic Hand Validity Check." To further improve the structural accuracy of the hand in 3D space, we apply a significantly higher weight ($\lambda_{z}=10$) specifically to the Z-axis (depth) component.

\noindent \textbf{Regularization Loss ($\mathcal{L}_{reg}$).}
To prevent the model from converging to biologically implausible states within an unconstrained optimization space, we apply $L_2$ regularization. This loss encourages the optimized pose parameters to remain close to their initial coarse estimates or pulls them toward canonical zero poses:
\begin{equation}
\mathcal{L}_{reg} = \lambda_1 \| \theta_{body} - \theta_{init} \|_2^2 + \lambda_2 \| \theta_{hands} - 0 \|_2^2 + \lambda_3 \|\beta\|_2^2 + \lambda_4 \|\psi\|_2^2
\end{equation}
Furthermore, we impose additional soft constraints on the jaw pose, overall human body scale, and global depth displacement (Z-axis vertices). For facial expression control, we introduce an extreme-value penalty featuring a truncation mechanism, formulated as $\text{ReLU}(|\psi| - 4.0)$. Whenever the absolute value of the expression coefficients exceeds the safety threshold of $4.0$, a substantial penalty is enforced. This strictly prevents exaggerated facial distortions and unnatural deformations during the tracking process.

\noindent \textbf{Temporal Smoothness Loss ($\mathcal{L}_{smooth}$).}
To eliminate the inherent frame-by-frame detection jitter from the driving video and ensure temporal coherence in the generated animation, we penalize the first-order derivatives (i.e., inter-frame differences) of the optimized parameters between adjacent frames ($i$ and $i-1$) using Mean Squared Error (MSE):
\begin{equation}
\begin{split}
    \mathcal{L}_{smooth} = \sum_{i=1}^{N} \bigg( & \lambda_{cam} \|\Delta C_{cam}\|_2^2 + \lambda_{pose} \|\Delta \theta\|_2^2 \\
    & + \lambda_{v} \|\Delta V_{body}\|_2^2 + \lambda_{exp} \|\Delta \psi\|_2^2 \bigg)
\end{split}
\end{equation}
Specifically, we assign stronger smoothness constraints to camera translation $\Delta C_{cam,T}$ and rotation $\Delta C_{cam,R}$ to simulate stable, real-world camera movements. Simultaneously, we suppress high-frequency fluctuations in hand poses, torso dynamics, and facial expressions.

\noindent \textbf{Optimization Details.}
The entire motion tracker is implemented in PyTorch~\cite{pytorch} and optimized using the AdamW optimizer over a total of 1,250 steps. To ensure efficient convergence, we employ distinct initial learning rates for different parameter groups: $10^{-3}$ for pose parameters, $5 \times 10^{-3}$ for camera translation, and $10^{-4}$ for identity-related parameters such as body shape.
To further improve convergence stability and final tracking quality, we apply learning rate decay at the 1,000th step. Concurrently, we adaptively readjust the loss weights by dynamically amplifying the temporal smoothness term ($\lambda_{smooth}$) while decaying the 3D alignment loss. Finally, before entering the optimization loop, all initial poses are preprocessed using a moving average filter to suppress high-frequency noise at the source.

\section{Further Experiments}
\label{spsec:more-results}

\subsection{Shape Disentanglement}
Without our disentanglement mechanism, the model suffers from shape leakage, causing the generated character's physique to improperly alter (e.g., from thin to robust), as shown in main paper \cref{fig:ablation}.
To quantitatively assess this, we utilized GVHMR to extract body shape parameters from the reference image and calculated their similarity against the estimated parameters from each frame of the generated video. As reported in \cref{sup_tab:shape_consistency}, our method yields substantial improvements in body shape consistency and disentanglement.

\begin{table}[tb]
\centering
\caption{ Evaluation of shape consistency.}
\resizebox{0.8\linewidth}{!}{
\begin{tabular}{lccc}
     \toprule
       &  Ours &  w/o Disentanglement & Wan-Animate    \\

    \midrule
   Shape Consistency Score $\uparrow$   & \bf 0.4483 & 0.3968 & 0.3560   \\
   \bottomrule
\end{tabular}
}

\label{sup_tab:shape_consistency}
\end{table}

\subsection{Identity Consistency.} 
\begin{figure}[tb]
  \centering
  \includegraphics[width=\linewidth]{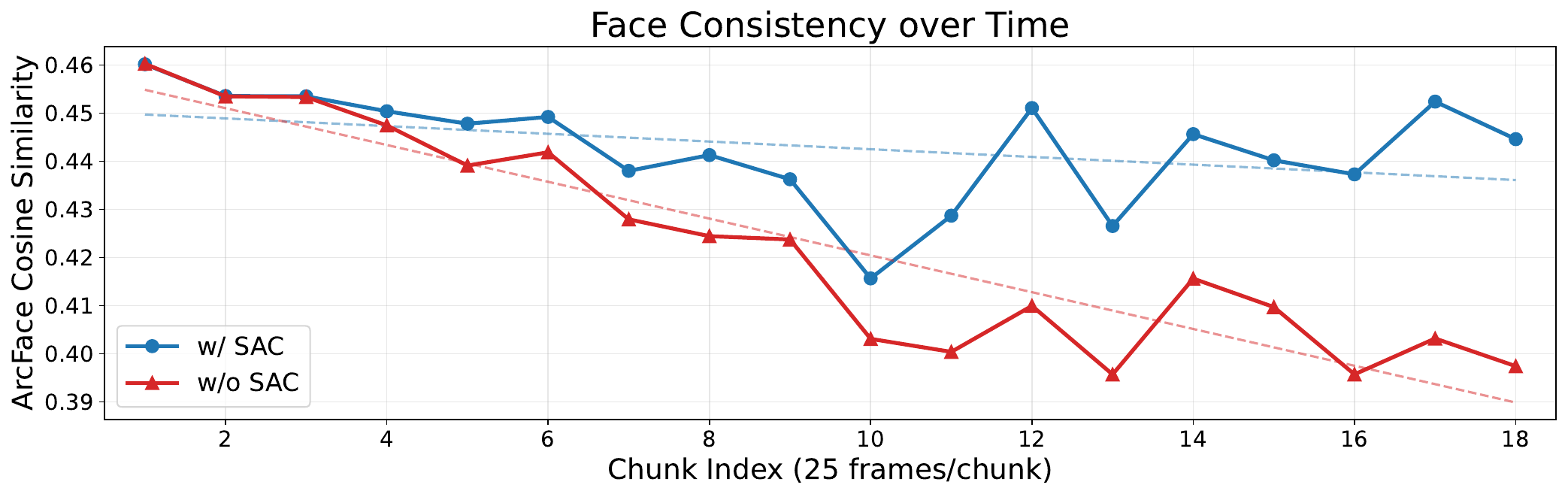}
   \caption{Long-term identity consistency with and without SAC.}
   \label{sup_fig:face_time_consistency}
\end{figure}

 We further explore how human identity consistency changes with the length of the generated video. As shown in \cref{sup_fig:face_time_consistency}, evaluating 120 generated videos over time reveals a widening performance gap between the "w/ SAC" and "w/o SAC" variants. Although varying video lengths cause minor statistical noise at the tail end, the increasing margin empirically indicates that our Spatially-Aligned
Conditioning (SAC) module effectively mitigates identity consistency degradation in long-term generation.

\begin{table}[tb]
\centering
\caption{Comparison of tracking efficiency on a 436-frame video.}
\label{sup_tab:tracking_efficiency}
\resizebox{0.8\linewidth}{!}{%
\begin{tabular}{lccc}
\toprule
\textbf{Method} & \textbf{Total Time (s)} $\downarrow$ & \textbf{Speed (FPS)} $\uparrow$ & \textbf{Relative Speedup} \\
\midrule
GUAVA Tracker & 424 & 1.03 & - \\
Our Tracker & \textbf{223} & \textbf{1.96} & \textbf{$\sim$1.9$\times$} \\
\bottomrule
\end{tabular}%
}
\end{table}

\begin{figure}[tb]
  \centering
  \includegraphics[width=\textwidth]{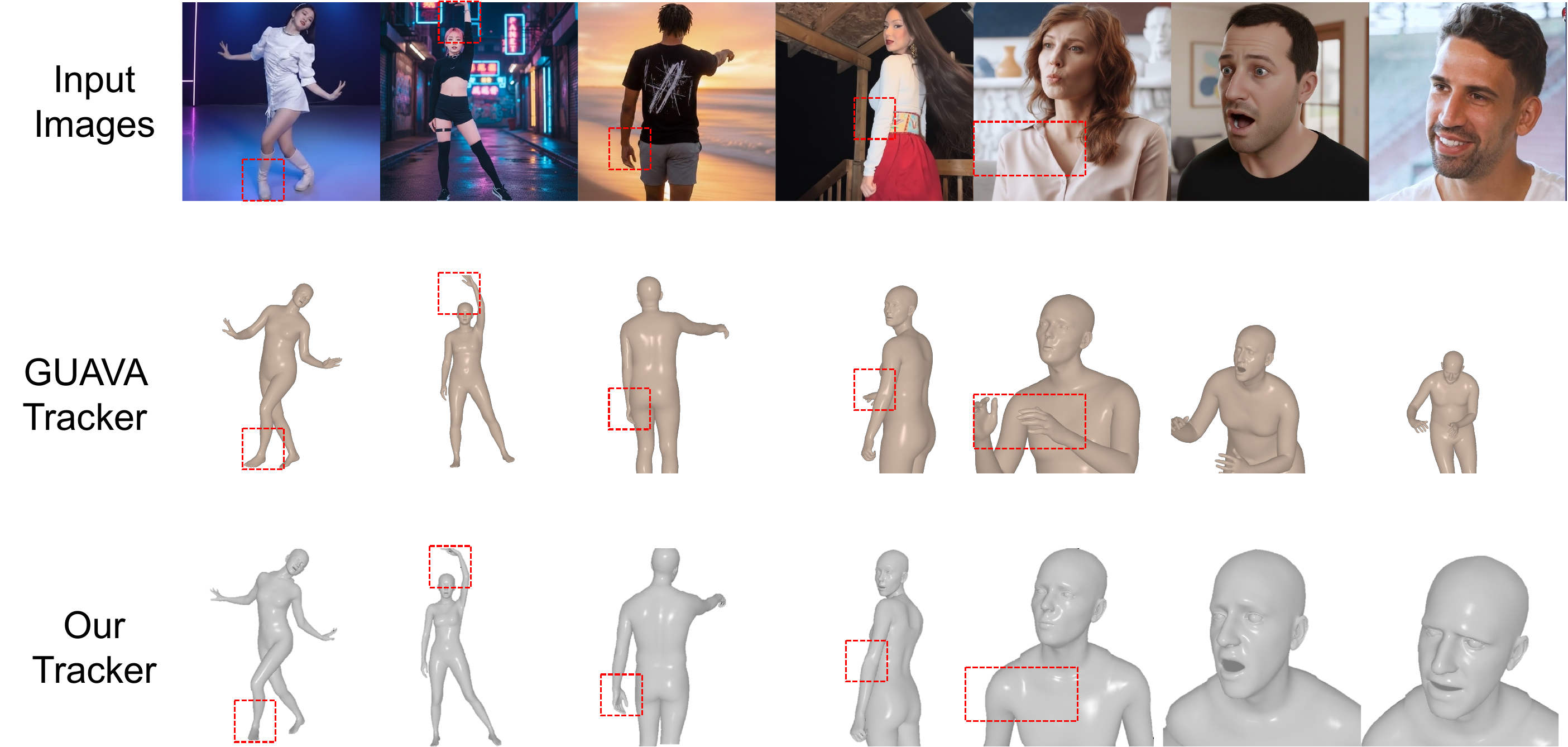}
 \caption{Visual comparison of tracking results. We compare our tracker with GUAVA across full-body, half-body, and head-only scenarios. As highlighted by the red dashed boxes, GUAVA is prone to foot/hand deviations, hallucinating occluded arms, and tracking collapse in extreme close-ups. Conversely, our method gracefully handles these complex scenes and severe occlusions, validating the effectiveness of our confidence-aware Validity Gating mechanism.}
  \label{supp_fig:compare_tracking}
\end{figure}

\subsection{Tracking Comparison}
Although we have demonstrated the robustness of our tracker across various poses and scenarios in the main paper \cref{fig:show_tracking}, we further compare it with the GUAVA tracker~\cite{guava} to underscore the superiority of our joint optimization framework and the Validity Gating mechanism.

As shown in~\cref{sup_tab:tracking_efficiency}, we evaluate tracking efficiency, a critical factor for processing large-scale video data. We test both methods on a 436-frame video. Our method requires approximately 223 seconds (around 1.96 FPS), whereas the GUAVA tracker takes about 424 seconds (around 1.03 FPS). This represents a relative speedup of nearly 90\%, significantly enhancing video tracking efficiency.

Beyond speed, tracking accuracy and robustness are of paramount importance. To this end, we compare the visual tracking results of both methods across diverse scenarios, as shown in~\cref{supp_fig:compare_tracking}. It is evident that in full-body scenarios, the GUAVA tracker exhibits substantial deviations in estimating foot and hand poses. In half-body scenarios where the arms are occluded (invisible), GUAVA erroneously hallucinates the arms into the visible frame. Furthermore, in head-only cases, the GUAVA tracker is prone to tracking collapse, resulting in severe estimation errors. In contrast, our method robustly and accurately estimates human poses across all these challenging conditions. This strongly validates the effectiveness of our Confidence-Aware Validity Gating mechanism in gracefully handling occlusions and complex, diverse scenes.

\begin{figure}[tb]
  \centering
  \includegraphics[width=\textwidth]{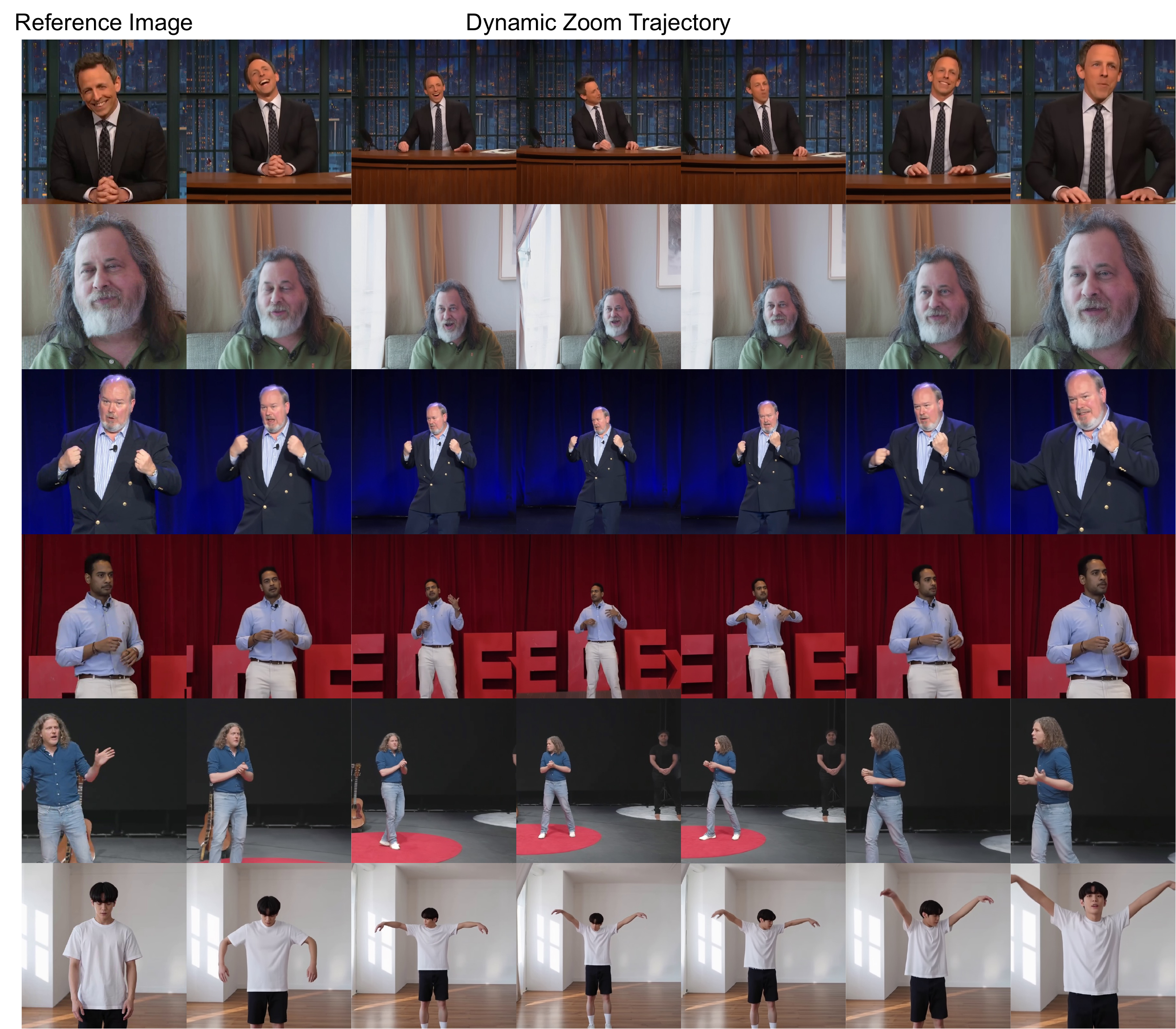}
\caption{Zero-shot dynamic zoom trajectory. We demonstrate the model's ability to perform simple camera movements (e.g., zoom-in and zoom-out) despite lacking explicit training on dynamic trajectories. By adjusting the rendering camera's distance, the resulting scale variation of the condition mesh acts as a spatial cue, implicitly guiding the model to follow the intended camera shifts.}
  \label{supp_fig:camera_zoom}
\end{figure}

\subsection{Camera Trajectory Control}
Although our model is \textbf{not} explicitly trained on datasets featuring dynamic camera trajectories, it exhibits a zero-shot capability to control simple camera movements, such as zoom-in and zoom-out effects. As illustrated in~\cref{supp_fig:camera_zoom}, by adjusting the distance of the rendering camera, the generated videos successfully follow the intended trajectory changes. This capability arises because modifying the camera distance alters the scale of the rendered mesh in the condition map. Serving as a visual reference, this scale variation implicitly enables the model to perceive the camera's spatial shift.

Unfortunately, our current framework struggles with rotational camera trajectories. Since we rely solely on the mesh as a spatial condition, the model cannot disambiguate between subject rotation and camera rotation. Because our training data predominantly consists of static cameras with moving or rotating subjects, rendering condition maps with a rotating camera merely forces the subject to rotate in the generated video, failing to produce the desired camera orbiting effect. Moving forward, incorporating video datasets with diverse camera trajectories and utilizing text annotations to explicitly decouple camera motion from human movement presents a promising direction for future improvement.

\subsection{Spatial Anchor Selection.}
\begin{figure}[tbh]
  \centering
  \includegraphics[width=\linewidth]{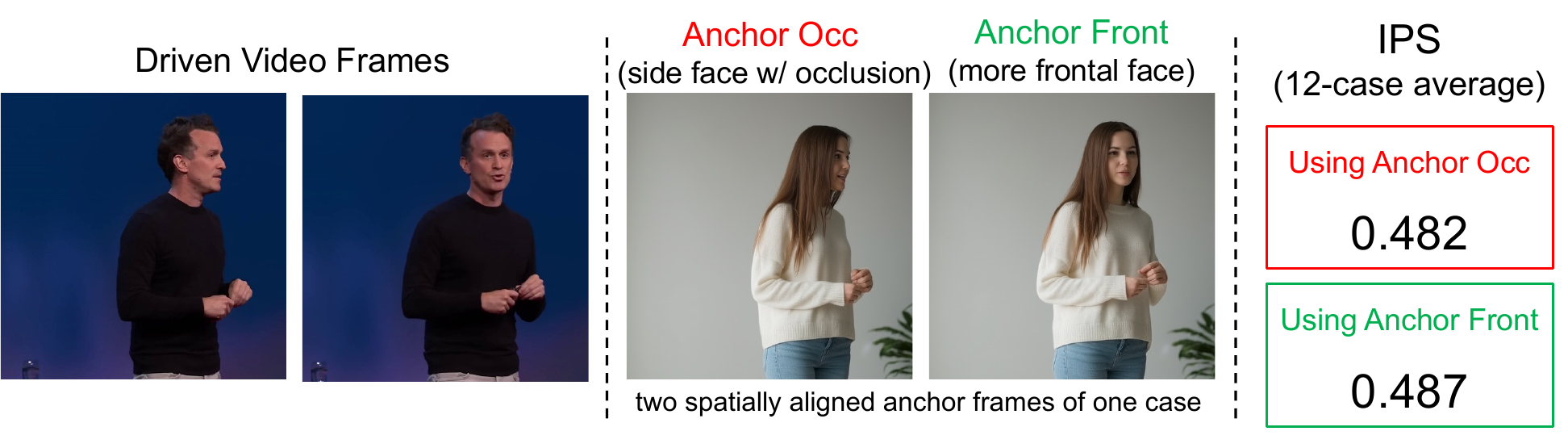}
   \caption{Impact of anchor frame selection on generation.}
   \label{supp_fig:sac_image_choosen}
\end{figure}

We evaluate two types of spatially aligned anchor frames, a partially occluded side view and a frontal view—across 12 cases, with results shown in \cref{supp_fig:sac_image_choosen}. Using a more frontal anchor yields a slight improvement in IPS, as expected; however, the performance gap is small, indicating robustness to moderate pose variation and occlusion.

\subsection{FID and FVD.}
We provide additional FID and FVD metrics evaluated on the EchoMimicV2 dataset in \cref{sup_tb:fid_fvd}. As shown in the table, our method achieves the lowest scores across both metrics, demonstrating its superiority in generating videos with higher spatial quality and better temporal consistency compared to other state-of-the-art baselines.

\begin{table}[tb]
    \centering
    \caption{FID and FVD metrics on EchoMimicV2 dataset.}
    \label{sup_tb:fid_fvd}
    \begin{tabular}{lcccc}
        \toprule
        & Ours & Wan-Animate & HyperMotion & UniAnimate-DiT \\
        \midrule
        FID $\downarrow$ & \textbf{25.86} & 27.91 & 33.94 & 112.02 \\
        FVD $\downarrow$ & \textbf{32.07} & 34.49 & 33.90 & 35.33 \\
        \bottomrule
    \end{tabular}
\end{table}
 
\section{More Discussion}
\label{spsec:more-discuss}
\subsection{Limitations}
Despite the promising results, our method still has several limitations. First, restricted by the quality and clarity of the available datasets, we currently only train a low-resolution version of EMOSH, which limits its broader application. Future work may explore scaling the model to higher resolutions, contingent on the availability of high-quality data. Second, due to the underlying architecture, the inference process is computationally demanding, requiring several minutes to generate a short video clip. Third, when tracking motions and camera trajectories from driving videos, relying solely on mesh conditioning leads to an inherent ambiguity between camera and human movement. Future work could incorporate text annotations to disentangle camera and human motions for finer-grained control. Fourth, during EHM retargeting, variations in body shapes across individuals can lead to spatial discrepancies. For instance, applying identical joint rotations to subjects with different proportions may alter the relative positions of end-effectors, occasionally misinterpreting a "clapping" gesture as "crossed hands." Finally, the generation quality is heavily contingent on the performance of the upstream motion tracker. Any tracking inaccuracies, such as mesh interpenetration, inevitably propagate to the final video, resulting in visual artifacts.

\subsection{Ethical Considerations}
Although EMOSH has made significant progress in generating high-fidelity human videos, we fully recognize the "dual-use" nature of such technologies and the potential ethical issues they may raise in real-world deployment.
First, our framework demonstrates robust capabilities in cross-driven video generation, driving static reference images with expressive motions. By achieving fine-grained control over facial expressions and hand gestures, the generated videos exhibit high realism and deceptiveness, which inevitably exacerbates the risk of Deepfakes. If misused, this technology could be weaponized for identity theft, financial fraud, or the dissemination of misinformation, thereby causing severe reputational damage to individuals. Second, the unauthorized collection and processing of sensitive biometric data, such as facial features and unique body shapes, raise stringent privacy concerns.
To mitigate these risks, we strongly advocate for responsible AI research and application. Users must strictly adhere to data protection regulations and ensure explicit informed consent is obtained prior to utilizing personal images. We explicitly oppose the unauthorized generation of identifiable public figures or private individuals. Looking ahead, we encourage the integration of robust synthetic media detection mechanisms, such as invisible watermarking and provenance tracking, into the video generation pipeline. This will enhance technical transparency, foster public trust, and effectively curb malicious abuse.

\end{document}